\pdfoutput=1

\documentclass[11pt]{article}

\usepackage[preprint]{acl}

\usepackage{times}
\usepackage{latexsym}

\usepackage[T1]{fontenc}

\usepackage[utf8]{inputenc}

\usepackage{microtype}
\usepackage{multirow}
\usepackage{booktabs, tabularx}
\usepackage{booktabs, amsmath}
\usepackage{pifont}
\usepackage{tikz}
\usepackage{bm}
\usepackage{inconsolata}
\usepackage{amssymb}

\newcommand*\mycirc[1]{%
  \begin{tikzpicture}
    \node[draw,circle,inner sep=1pt] {#1};
  \end{tikzpicture}}
  
\usepackage{inconsolata}

\usepackage{graphicx}

\title{NormXLogit: The \textit{Head-on-Top} Never Lies}

\author{
    Sina Abbasi\textsuperscript{1}
    ~ Mohammad Reza Modarres\textsuperscript{1}
    ~ Mohammad Taher Pilehvar\textsuperscript{2} \\
    \textsuperscript{1} Tehran Institute for Advanced Studies, Khatam University, Iran \\
    \textsuperscript{2} Cardiff University, United Kingdom\\
    \texttt{s.abbasi401@khatam.ac.ir} ~
    \texttt{m.modares401@khatam.ac.ir} ~
    \texttt{mp792@cam.ac.uk}
}

\begin{document}
\maketitle
\begin{abstract}
With new large language models (LLMs) emerging frequently, it is important to consider the potential value of model-agnostic approaches that can provide interpretability across a variety of architectures.
While recent advances in LLM interpretability show promise, many rely on complex, model-specific methods with high computational costs.
To address these limitations, we propose NormXLogit, a novel technique for assessing the significance of individual input tokens.
This method operates based on the input and output representations associated with each token.
First, we demonstrate that during the pre-training of LLMs, the norms of word embeddings effectively capture token importance.
Second, we reveal a significant relationship between a token's importance and the extent to which its representation can resemble the model's final prediction.
Extensive analyses reveal that our approach outperforms existing gradient-based methods in terms of faithfulness and offers competitive performance in layer-wise explanations compared to leading architecture-specific techniques.
\end{abstract}

\section{Introduction}

Transformer-based models have gained widespread adoption across various natural language processing (NLP) tasks, demonstrating their versatility.
However, the underlying mechanisms of these models are not quite understood.
This means when the model fails and generates inaccurate or harmful content, we are unable to diagnose the source and improve the model's behavior.
Consequently, a multitude of endeavors in recent years aimed at enhancing the interpretability of these models.

Architecture-agnostic methods, such as perturbation-based and gradient-based techniques, are widely used to identify influential input tokens that impact a model's predictions. However, these approaches face significant challenges, including the generation of out-of-distribution inputs and vulnerability to adversarial exploitation \citep{wang-etal-2020-gradient}.
Additionally, they often come with high computational costs, limiting their scalability.
For instance, advanced gradient-based methods like Integrated Gradients \citep{pmlr-v70-sundararajan17a} involve repeated forward and backward passes, leading to substantial overhead.

\begin{figure}
    \centering
    \includegraphics[width= \linewidth]{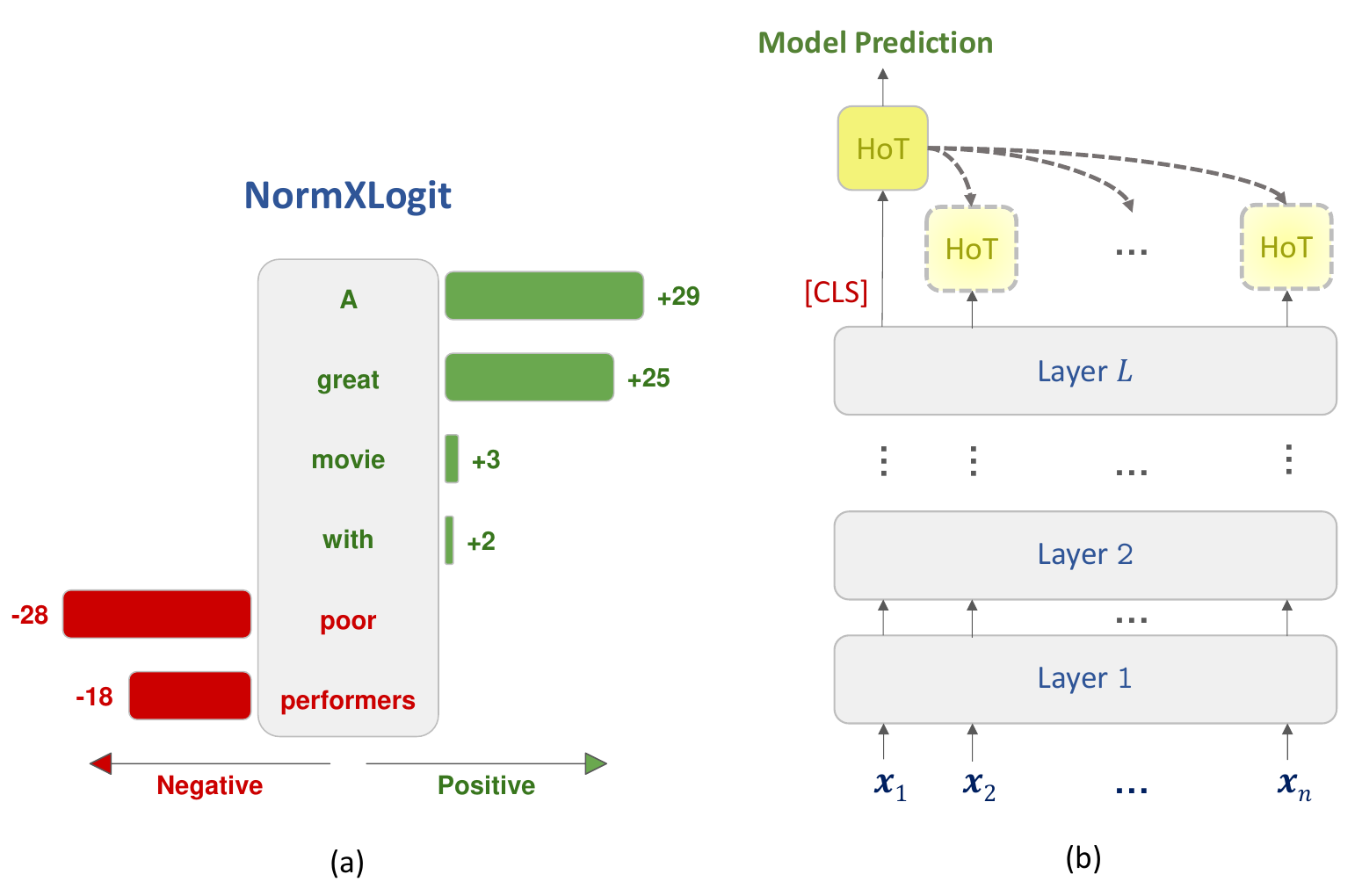}
    \caption{(a) Importance scores of NormXLogit for the sentiment analysis task. NormXLogit generates attributions per-label with signed scores denoting positive/negative impact. (b) Applying the head-on-top (HoT) on each of the final representations to obtain a prediction based on each token.}
    \label{head-on-top}
\end{figure}

By leveraging the internal components of the target model, a new class of architecture-specific approaches, termed vector-based methods, has been developed
(\citealp[]{kobayashi-etal-2021-incorporating, ferrando-etal-2022-measuring}).
Most of these approaches offer per-layer explanations, which are subsequently aggregated using the \textit{rollout} technique \citep{abnar-zuidema-2020-quantifying} to achieve a global interpretation that integrates all layers of the model.
However, rollout can result in inaccurate outcomes due to the vanishing attribution problem \citep{Mehri_2024_CVPR}.
Despite recent advances in faithfulness, a key drawback of vector-based methods is their architecture-specific design, which limits their adaptability to the rapidly evolving landscape of LLMs.
Furthermore, almost all of these methods ignore the \textit{head-on-top}\footnote{By `head-on-top', we are referring to the classification or regression head used on top of the pre-trained models.} which is crucial to produce task-dependent explanations.

In response to these limitations, this paper presents NormXLogit, a simple yet powerful, computationally efficient, and architecture-agnostic\footnote{By using the term "architecture-agnostic," we aim to differentiate our approach from vector-based methods, ensuring generalization across different Transformer-based models.} approach that outperforms many sophisticated techniques and can be easily applied to any NLP task.
NormXLogit eliminates the need for any backward passes, utilizing the pre-trained model only once.
It leverages the informativeness of the norm of input embeddings, in conjunction with the rich semantic and syntactic information encoded in the model’s final-layer representations. The latter enables the incorporation of all internal components, thereby removing the need for any aggregation method, such as rollout.
We incorporate the head-on-top of pre-trained models into our analysis, which serves as a task-specific interpreter of token attributions\footnote{We use `attribution' and `importance' interchangeably.}, tailoring its attention to the nuances of the task.
This helps NormXLogit to generate per-label attributions with positive and negative impacts of each input token (cf. Figure \ref{head-on-top}(a)).

Based on comprehensive experiments across multiple tasks and models, we show that in numerous instances, the faithfulness of NormXLogit surpasses that of widely recognized gradient-based approaches.
In the regression setup, it demonstrates superior performance compared to a state-of-the-art architecture-specific baseline.
Furthermore, through a targeted experiment on the BLiMP corpus \citep{warstadt-etal-2020-blimp}, we evaluate the alignment between NormXLogit's attributions and known linguistic evidence across diverse grammatical phenomena. Our results demonstrate that NormXLogit consistently yields more faithful and plausible explanations than existing approaches, highlighting its effectiveness as a general-purpose token attribution method.

\section{Related Work}

In recent years, vector-based analyses have emerged as an architecture-specific approach to Transformer interpretability. These methods build on findings that attention weights can be misleading \citep[]{jain-wallace-2019-attention, serrano-smith-2019-attention}.
\citet{kobayashi-etal-2021-incorporating} extended analysis beyond attention weights by decomposing the entire attention block. To aggregate per-layer attributions into global ones, \citet{abnar-zuidema-2020-quantifying} proposed \textit{attention rollout} to quantify the flow of information through self-attention.

GlobEnc \citep{modarressi-etal-2022-globenc} and ALTI \citep{ferrando-etal-2022-measuring} advanced previous work by further incorporating other Transformer components.
ALTI challenged the use of $\ell^2$ norms for measuring contributions, proposing the Manhattan distance for improved results.
GlobEnc added the second residual connection and layer normalization into the analysis but overlooked the impact of the feed-forward network.
To address this, DecompX \citep{modarressi-etal-2023-decompx} captured the influence of the feed-forward network by approximately decomposing the activation function and propagating decomposed vectors across layers, eliminating reliance on aggregation methods like rollout.

Perturbation-based methods, such as SHAP \citep{lundberg2017unified} and LIME \citep{ribeiro2016why}, investigate the causal link between input features and the model's final prediction by perturbing or erasing parts of the input.
Recently, \citet{mohebbi-etal-2023-quantifying} proposed \textit{Value Zeroing} which is based on the Explaining-by-Removing intuition \citep{covert2022explaining}, in order to quantify the context mixing. Their approach, in contrast to other perturbation-based methods, does not remove the input token representations. Value Zeroing instead suggests zeroing the value vector of each token to measure its contribution.

Gradient-based methods involve analyzing the gradients of the model's output with respect to the input features to understand their impact on the model's decision-making process. Methods such as Gradient Norm \citep{simonyan2014deep}, Gradient$\times$Input \citep{kindermans2016investigating}, and Integrated Gradients \citep{pmlr-v70-sundararajan17a} are the most prominent ones in this category.

\section{Proposed Approach}

\subsection{Background: Transformer Architecture}
The Transformer architecture consists of multiple identical layers, each with a multi-head self-attention (MHA) block and a position-wise fully connected feed-forward network (FFN), where both are followed by a residual connection (RES) and layer normalization (LN).

The MHA component is responsible for creating contextualized representations for the input elements.
The output representation of MHA for token $x_i$ and head $h$ in $l$-th layer is then calculated as the weighted sum of transformed input representations:
\begin{equation} \label{eq7}
\Tilde{x}_i^{l \hspace{0.75mm} \scriptsize{\mycirc{h}}} = \sum_{j=1}^n \alpha_{i, j}^{l \hspace{0.75mm} \scriptsize{\mycirc{h}}} v_j^{l \hspace{0.75mm} \scriptsize{\mycirc{h}}}
\end{equation}
where $\alpha_{i, j}^{l \hspace{0.75mm} \scriptsize{\mycirc{h}}}$ represents the attention weight of token $i$ with respect to token $j$ in the $h$-th head of the MHA of the $l$-th layer.

\citet{kobayashi-etal-2021-incorporating} demonstrated that the output representation of each token produced by the attention block can be explained via two effects: (i) “preserving” its original input using the RES and the contribution of the token itself through context mixing of MHA, and (ii) “mixing” the representations in the context (except the target token).
They showed that the preserving effect is predominant, primarily due to the higher contribution of RES to the output representation.

\subsection{Norm of Word Embedding}
\citet{oyama-etal-2023-norm} demonstrated that the norm of input embeddings encodes information gain.
They showed that tokens with higher $\ell^2$ norm carry more information, effectively capturing the least frequent words in the text.
Additionally, based on the Eq. \ref{eq7}, the MHA could be interpreted as the weighted sum of transformed vectors.
In other words, the final representation of each token is built by mixing the representations of all tokens in the input sequence. Consequently, tokens with higher norms are expected to contribute more to the final representation of the target token. Higher contribution suggests greater importance, allowing us to utilize the $\ell^2$ norm of word embeddings to identify crucial tokens influencing the model's decision.

\subsection{LogAt: Logit Attribution}
\label{sec:LogAt}

The tasks in the domain of NLP can be broadly divided into two main categories: classification tasks and regression tasks.
For both of these setups, we utilize a special token (often known as \texttt{[CLS]}\footnote{The name of this special token may vary depending on the model. Also, in auto-regressive models, the last token in the input is typically used for classification.}), which is embedded in almost all pre-trained models.
This token serves as a single vector representing the entire input sequence, which is then fed into head-on-top, an FFN placed on top of the pre-trained model to produce the output prediction.

The intuition behind the attention mechanism implies that more important tokens have a greater contribution to building the final representation of the \texttt{[CLS]} token.
In other words, an identical \texttt{[CLS]} embedding is fed into the model for all input sequences, and based on the fine-tuning objective, the attention block attempts to utilize the most relevant (i.e., important) tokens to construct the new representation of \texttt{[CLS]}.
This suggests that the \texttt{[CLS]} token has a higher degree of similarity to the most important input tokens in the model's decision-making process.
To identify the tokens that are most similar to \texttt{[CLS]}, we use the head-on-top to evaluate how each individual token in the input contributes to predicting the target task.
In the following, we describe the approach for each setup.

\paragraph{Classification.} 
In a classification setup, the output of the head-on-top for each sample is a vector of length equal to the number of labels (i.e., classes).
The values in this output vector are referred to as logits, which are further processed using the softmax function to obtain probabilities over the output labels.
The model’s final prediction is the label associated with the highest logit value.
To determine the most important input tokens for a model with $L$ layers, we apply the head-on-top to each one of the output representations at layer $L$, as illustrated in Figure \ref{head-on-top}(b).
Next, we extract the logits corresponding to the predicted class, which is already determined by applying the head-on-top to the \texttt{[CLS]} token.
The logit value associated with each token represents its attribution, and tokens with the highest logits are regarded as the most important for the classification task.
We call this method Logit Attribution (LogAt).
To calculate the attribution ($\text{Att}_\text{LogAt}$) of the token $i$ for a task with $C$ classes and a classification head $\text{HoT}_\text{clas}(\cdot)$ $\in \mathbb{R}^C$, we have:
\begin{equation} \label{classification FFN}
\text{Att}_\text{LogAt}(x_i) = \text{HoT}_\text{clas}(x_i^L)[\hat{p}]
\end{equation}
where $x_i^L$ is the final representation of the $i$-th token in a model with $L$ layers, and $\hat{p} \in \{ 0, 1, ..., C-1\}$ denotes the index of the predicted class.
By changing the index of $\hat{p}$ to other class labels in the task, we can identify the important tokens relative to those classes as well, leading to a per-label attribution technique.

Due to the dominance of the “preserving” effect in the attention block, the contextualized representations in the last layer still retain the identity of the original input tokens.
As a result, the logits can be seen as a direct reflection of each token's contribution.
The use of the head-on-top provides task-specific explanations, allowing us to semantically identify the tokens that are most critical for the target task.
Furthermore, the sign of the logits provides insight into the direction of the contributions, indicating whether each label is positively or negatively influenced, specifically with respect to the model's predicted label.

To interpret the choice of token in various \textbf{language modeling objectives}, we categorize them as classification tasks, where the number of labels corresponds to the vocabulary size.
In language modeling, the goal is to predict the correct word given the context, which yields a probability distribution over the vocabulary for generating each individual word.
In this setup, we utilize the masked language modeling head as the head-on-top to identify the tokens that contribute most to predicting the \texttt{[Mask]} token.

\paragraph{Regression.}
For the regression setup, the approach typically involves generating a single value in the output rather than a vector of probabilities.
So, instead of taking the largest logit corresponding to the prediction label, we take the absolute distance of the output for each token from the model's prediction.
For the attribution of $i$-th token in a regression task, we have:
\begin{equation} \label{regression att}
\text{Att}_\text{LogAt}(x_i) = \left\lvert \text{HoT}_\text{reg}(x_i^L) - \text{HoT}_\text{reg}(\texttt{[CLS]}^L)\right\rvert
\end{equation}
where $\text{HoT}_\text{reg}(\cdot) \in \mathbb{R}$ denotes the regression head, $x_i^L$ is the final representation of the $i$-th token in a model with $L$ layers, and $\texttt{[CLS]}^L$ is the final representation of the \texttt{[CLS]} token.

\subsection{NormXLogit}
Although LogAt provides valuable explanations of the model's decision-making process, our experiments show that considering the informativeness of the norm of word embeddings can yield more faithful results.
Therefore, we introduce NormXLogit, an architecture-agnostic interpretation method that can be applied to any task and domain.
The attribution of token $i$ using NormXLogit is obtained as:
\begin{equation} \label{NormXLogit}
\text{Att}_\text{NormXLogit}(x_i) = \left\lVert x_i^0 \right\rVert_2 \times \text{Att}_\text{LogAt}(x_i)
\end{equation}
where the $\left\lVert x_i^0 \right\rVert_2$ is the $\ell^2$ norm of the input word embedding for the $i$-th token, and $\text{Att}_\text{LogAt}(\cdot)$ is the LogAt attribution according to the task setup.

\section{Experiment 1: Faithfulness Analysis}
\label{sec:experiment1}
To analyze the \textit{faithfulness} of NormXLogit, we begin with a set of experiments on classic classification and regression tasks.
The concept is straightforward: if a token is identified as highly important, altering or removing it should significantly impact the model's output prediction.
A faithful attribution method should effectively identify these critical tokens, leading to a more substantial drop in the model’s performance.

\subsection{Experimental Setup}
\paragraph{Data.}
In the classification setup, we will use SST-2 \citep{socher-etal-2013-recursive} for sentiment analysis, MultiNLI \citep{williams-etal-2018-broad} for recognizing textual entailment, and QNLI for question answering \citep{wang-etal-2018-glue}.
SST-2 contains movie review sentences labeled as positive or negative, while MultiNLI includes sentence pairs labeled as entailment, contradiction, or neutral. QNLI consists of question–sentence pairs labeled for entailment. STS-B \citep{cer-etal-2017-semeval} is used for semantic similarity as a regression task, with scores from 0 (no similarity) to 5 (semantic equivalence).

\paragraph{Models.}
Our target models in this section involve three prominent models: \textsc{Llama 2} \citep{Touvron2023Llama2O}, DeBERTa \citep{he2023debertav3}, and BERT \citep{devlin-etal-2019-bert}.\footnote{We employ the 7 billion parameter variant of \textsc{Llama 2}, the uncased BERT\textsubscript{base} model, and DeBERTaV3\textsubscript{base}, all from HuggingFace's Transformers library \cite{wolf-etal-2020-transformers}.}
We use the fine-tuned version of each model for the corresponding task.
To fine-tune \textsc{Llama 2} and perform inference, we employ the LoRA \citep{hu2021lora} technique with a rank of 4 from the PEFT library\footnote{\url{https://github.com/huggingface/peft}}.

\paragraph{Input Attribution Methods.}
To analyze the performance of our proposed method, we compare NormXLogit with three well-known gradient-based input attribution methods: Gradient Norm (Grad. Norm), Gradient$\times$Input (G$\times$I), and Integrated Gradients (IG), using the $\ell^1$ norm for aggregation.
Notably, we focus on gradient-based methods over perturbation-based approaches due to their more faithful results.
To account for vector-based approaches, we adopt DecompX as the current state-of-the-art method among them.
However, this family of methods is primarily developed for BERT-like architectures and may not be applicable to all models.
In addition, we consider a random baseline where tokens are ranked randomly from most important to least important.

\paragraph{Evaluation Metrics.}
To assess the faithfulness of the aforementioned methods, we utilize two metrics: AOPC\footnote{AOPC stands for ``Area Over the
Perturbation Curve."} \citep{DBLP:journals/corr/SamekBMBM15} for classification tasks and Accuracy for regression setups.

\paragraph{AOPC:}
This metric involves perturbing $K\%$ of the most important tokens in the input sequence and observing the resulting changes in the model's predictions.
For the masked language modeling objectives, masking is used for token perturbations, while for auto-regressive models, deletions are employed due to the absence of a \texttt{[MASK]} token.
For a given input sequence $X_i$, the perturbed sequence $X_i\backslash K$ is generated by applying the perturbation on $K\%$ of the most important tokens.
Then, for all of the instances in the dataset, the average AOPC is defined as:
\begin{equation} \label{eq3}
\text{AOPC}(K\%) = \frac{1}{m} \sum_{i = 1}^{m} \bigl[ f_{\hat{y}}(X_i) - f_{\hat{y}}(X_i\backslash K) \bigr]
\end{equation}
where $m$ is the number of instances, and $f_{\hat{y}}(X)$ is the model's output probability for label $\hat{y}$.
A higher AOPC for the predicted class indicates that the model exhibits a larger drop in its probability, reflecting greater sensitivity to perturbed tokens.

\begin{figure}[t]
    \centering
    \includegraphics[width= \linewidth]{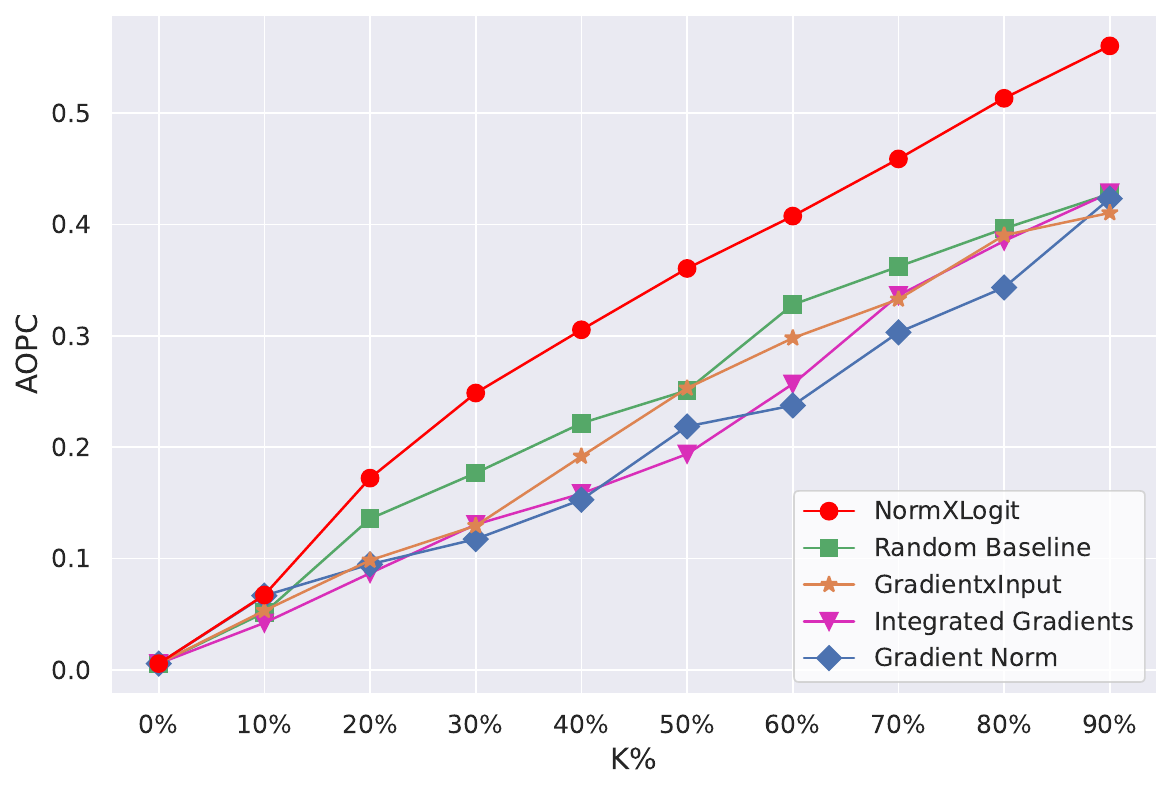}
    \caption{AOPC of different attribution methods for \textsc{Llama 2} fine-tuned on SST-2 (higher AOPC is better).}
    \label{SST2-Llama2-Norm2}
\end{figure}

\paragraph{Accuracy:} This metric evaluates model performance when different proportions (e.g., $10\%$, $20\%$, ...) of the most important input tokens are perturbed.
For regression tasks, we utilize the Pearson correlation coefficient as the Accuracy metric.
For classification tasks, we report standard accuracy (i.e., the percentage of correctly predicted labels) on the perturbed inputs.
In both cases, lower values indicate better attribution performance, as the model is more affected by key perturbations.

\subsection{Results}
Figure \ref{SST2-Llama2-Norm2} illustrates the superior performance of NormXLogit in \textsc{Llama 2} fine-tuned on SST-2.
NormXLogit achieves higher AOPC scores across all thresholds, indicating its effectiveness in identifying crucial tokens for the model's decision-making process.
It should be noted that DecompX, due to its architecture-specific nature, may not be applicable to \textsc{Llama 2}.
Additionally, even if it were, the computational cost of DecompX may not be easily manageable given the size of \textsc{Llama 2} on many accessible hardware configurations.

\begin{figure}[t]
    \centering
    \includegraphics[width= \linewidth]{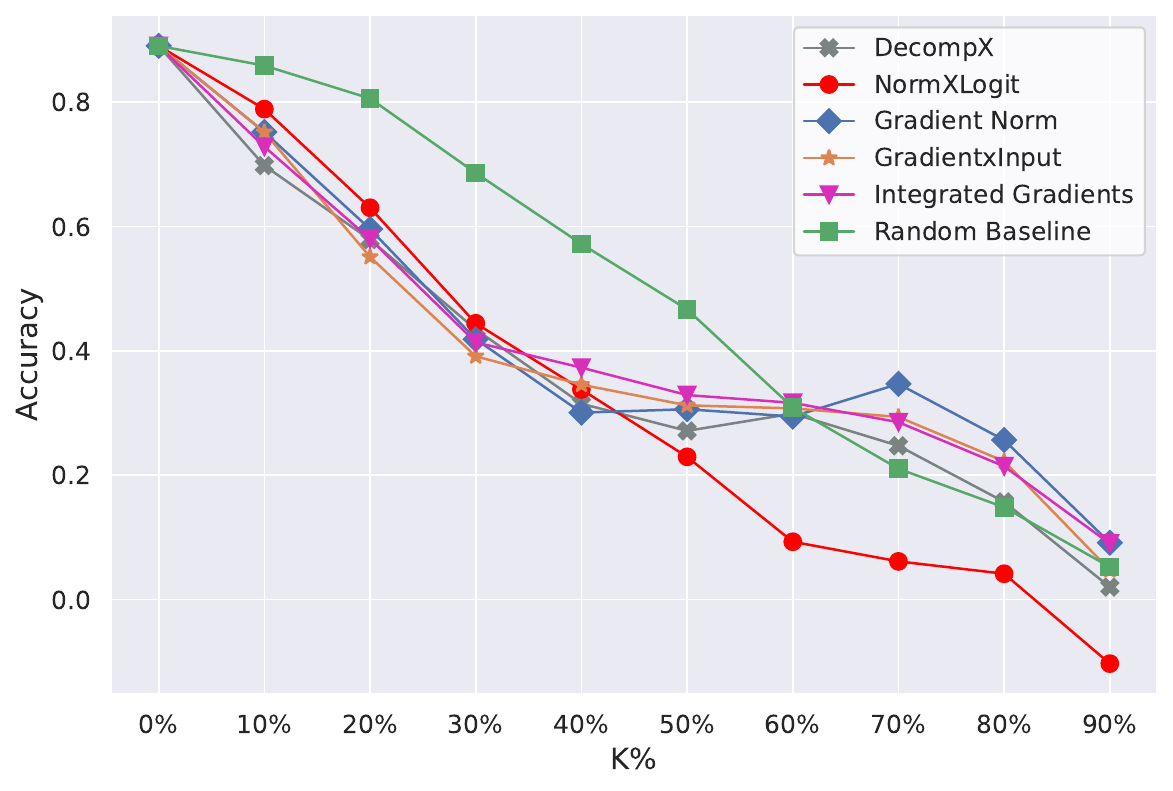}
    \caption{Accuracy of different attribution methods for BERT fine-tuned on STS-B (lower Accuracy is better).}
    \label{STS-B-BERT-Norm2}
\end{figure}

In the regression setup of STS-B depicted in Figure \ref{STS-B-BERT-Norm2}, dropping important tokens results in a decrease in Accuracy.
To leverage DecompX for this setup, in the absence of a classification head, we applied the $\ell^2$ norm to the decomposed vectors obtained from the final layer.
The results for the initial $K\%$ ratios are very close, with DecompX and Gradient$\times$Input showing a slight lead at the outset.
However, after dropping $40\%$ of the most important tokens, the performance of all these methods deteriorates, while NormXLogit continues to experience a drop in Accuracy.

\begin{table*}[h]
\begin{center}
\small
\tabcolsep=0.12cm
\renewcommand{\arraystretch}{1.04}
\begin{tabular}{l c c c | c c c | c c c | c c c} 
 \toprule
     & 
     \multicolumn{3}{c}{\textbf{\textsc{SST-2}} \scriptsize{\text{\textsc{(AOPC$\uparrow$)}}}} & 
     \multicolumn{3}{c}{\textbf{\textsc{MNLI}} \scriptsize{\text{\textsc{(AOPC$\uparrow$)}}}} & 
     \multicolumn{3}{c}{\textbf{\textsc{QNLI}} \scriptsize{\text{\textsc{(AOPC$\uparrow$)}}}} & 
     \multicolumn{3}{c}{\textbf{\textsc{STS-B}} \scriptsize{\text{\textsc{(Acc$\downarrow$)}}}} \\
     \cmidrule(lr){2-13}

    & \scriptsize{\textbf{\textsc{Llama 2}}} & \scriptsize{\textbf{DeBERTa}} & \scriptsize{\textbf{BERT}}
     & \scriptsize{\textbf{\textsc{Llama 2}}} & \scriptsize{\textbf{DeBERTa}} & \scriptsize{\textbf{BERT}}
     & \scriptsize{\textbf{\textsc{Llama 2}}} & \scriptsize{\textbf{DeBERTa}} & \scriptsize{\textbf{BERT}}
     & \scriptsize{\textbf{\textsc{Llama 2}}} & \scriptsize{\textbf{DeBERTa}} & \scriptsize{\textbf{BERT}} \\
    \midrule
    Random
& 0.256 & 0.266 & 0.245 & 0.421 & 0.445 & 0.361 & 0.284 & 0.306 & 0.273 & 0.283 & 0.430 & 0.457 \\
    Grad. Norm
& 0.216 & 0.320 & 0.331 & 0.419 & 0.535 & 0.460 & 0.334 & 0.365 & 0.360 & 0.351 & 0.338 & 0.374 \\
    G$\times$I
& 0.236 & 0.345 & 0.339 & 0.442 & 0.565 & 0.456 & 0.353 & 0.382 & 0.364 & 0.255 & \underline{0.214} & 0.358 \\
    IG
& 0.220 & 0.346 & 0.367 & 0.448 & \textbf{0.571} & 0.466 & 0.336 & 0.381 & 0.364 & 0.237 & 0.227 & 0.370 \\
    DecompX
& N/A & N/A & \textbf{0.574} & N/A & N/A & \textbf{0.585} & N/A & N/A & \textbf{0.460} & N/A & N/A & 0.336 \\
    \midrule
    $\ell^2$ norm
& 0.299 & 0.360 & 0.311 & 0.420 & 0.473 & 0.393 & 0.272 & 0.339 & 0.304 & 0.251 & \textbf{0.199} & 0.321 \\
LogAt
& \underline{0.341} & \underline{0.377} & 0.364 & \underline{0.518} & 0.548 & \underline{0.566} & \textbf{0.378} & \underline{0.435} & 0.394 & \textbf{0.167} & 0.423 & \underline{0.313} \\
    \textbf{NormXLogit}
& \textbf{0.341} & \textbf{0.386} & \underline{0.423} & \textbf{0.519} & \underline{0.566} & 0.556 & \underline{0.363} & \textbf{0.474} & \underline{0.402} & \underline{0.233} & 0.320 & \textbf{0.281} \\
 \bottomrule
\end{tabular}
\end{center}
\caption{
Performance evaluation of NormXLogit against other methods across various model and dataset configurations.
Each value is computed by averaging across all perturbation ratios (higher AOPC and lower Accuracy are better). Best values are in \textbf{bold}, and second-best values are \underline{underlined}.
}
\label{AOPC-Acc}
\end{table*}

Table \ref{AOPC-Acc} presents the average AOPC and Accuracy across different ratios of perturbation, evaluated on various models and datasets\footnote{The corresponding diagrams are provided in Section \ref{sec:experiment1_apx}.}.
For classification tasks, the Accuracy metric results are consistent with AOPC scores and are provided in the appendix to avoid repetition (cf. Table \ref{Tab:Acc}).
In SST-2 and QNLI, NormXLogit consistently outperforms architecture-agnostic methods.
However, DecompX, which is specific to the BERT architecture, results in a higher drop in AOPC.
In the MultiNLI dataset, NormXLogit performs better than gradient-based approaches in \textsc{Llama 2} and BERT, though Integrated Gradients show a slight edge in the DeBERTa model.
In the BERT model, similar to SST-2, DecompX performs better but the difference is notably smaller compared to the SST-2 dataset.

In the regression setup, the $\ell^2$ norm performs surprisingly well and also boosts the performance of Gradient$\times$Input.
This strong performance on STS-B can be attributed to the structure of STS-B samples:
sentence pairs consist of relatively short sentences with similar openings, where key information that determines the label often resides in less frequent words toward the end.
Such words tend to have higher embedding norms, making the $\ell^2$ norm approach more effective.

Furthermore, in the BERT model, the absence of a classification head diminishes DecompX’s effectiveness, resulting in performance worse than that of the input embeddings' norms.
In \textsc{Llama 2}, NormXLogit slightly surpasses all gradient-based methods, largely due to the strong performance of LogAt.

\paragraph{Computational Efficiency.}
NormXLogit is not only faithful and generalizable, but also computationally lightweight.
In perturbation-based methods, it is costly to search for appropriate combinations of tokens to intervene, as this requires multiple forward passes.
Similarly, advanced gradient-based methods like Integrated Gradients incur substantial overhead due to repeated forward and backward passes.
In contrast, NormXLogit performs attribution in a single forward pass, eliminating the need for any backpropagation.
Our runtime analysis shows that for longer inputs (e.g., 360 tokens or more), NormXLogit can be up to $6\times$ faster and up to $750\times$ more memory-efficient than its counterparts (cf. Tables~\ref{Tab:time} and~\ref{Tab:memory}).

\section{Experiment 2: Evidence Alignment}
In our second experiment, we extend our evaluation beyond classification and regression by focusing on the language modeling task, using the BLiMP dataset \cite{warstadt-etal-2020-blimp}. This experiment serves four purposes: (1) to assess the \textit{faithfulness} of NormXLogit in the masked language modeling setting, expanding the scope of our evaluation, (2) to examine its \textit{plausibility}, i.e., the extent to which its attributions align with human-understood linguistic rationales provided by BLiMP, (3) to analyze the target model’s sensitivity to specific linguistic phenomena, and (4) to investigate how \textit{contextualization} evolves throughout the model by analyzing the output representations at different layers of the Transformer (i.e., per-layer attributions).

\subsection{Experimental Setup}

\paragraph{Data.}
The BLiMP dataset contains sentence pairs with minimal contrasts in syntax, morphology, or semantics.
It is constructed to provide samples where the true label is uniquely determined by a single word in each sentence, providing an ideal benchmark for assessing attribution methods.
This word, which serves as the decisive factor in determining grammatical acceptability, is termed the \textit{evidence}.
BLiMP offers a strong prior on which token is expected to drive the model’s prediction, thereby supporting a targeted assessment of attribution faithfulness.
Additionally, it enables evaluation of plausibility, as the evidence in this dataset inherently aligns with human reasoning.
In summary, using BLiMP we are evaluating both the faithfulness and plausibility of these methods.

\begin{table}[t]
\centering
\setlength{\tabcolsep}{2.3pt}
\resizebox{\linewidth}{!}{ 
\begin{tabular}{lcl}
\toprule
\textbf{Phenomenon} & \textbf{UID}  & \textbf{Example} (Target \text{\ding{52}}/Foil \text{\ding{56}})\\ 
\midrule
\multirow{1}{*}{Anaphor Number Agreement} 
& ana & \underline{This government} alarms itself \text{\ding{52}}/themselves \text{\ding{56}}.\\
\midrule
\multirow{2}{*}{Determiner-Noun Agreement} & dna & Russell explored this \text{\ding{52}}/these \text{\ding{56}} \underline{mall}.\\
& dnaa & Patients scan this \text{\ding{52}}/these \text{\ding{56}} orange \underline{brochure}.\\
\midrule
\multirow{2}{*}{Subject-Verb Agreement} & darn & The \underline{sister} of doctors writes \text{\ding{52}}/write \text{\ding{56}}.\\
& rpsv & The \underline{pedestrian} has \text{\ding{52}}/have \text{\ding{56}} forgotten Grace. \\
\bottomrule
\end{tabular}}
\caption{Examples of various linguistic phenomena that have been investigated in our experiments. Each paradigm is represented by a unique identifier (UID) from the BLiMP dataset. The target and foil words are denoted using check and cross marks. In each instance, the relevant evidence is \underline{underlined}.} 
\label{blimp-table}
\end{table}

Following \citet{mohebbi-etal-2023-quantifying}, we utilize a subset of the BLiMP dataset comprising five paradigms that represent three distinct linguistic phenomena.
Examples of each phenomenon are provided in Table~\ref{blimp-table}.
Using spaCy \citep{spacy2}, we are able to identify the evidence of each linguistic phenomenon.
For \textit{anaphor number agreement}, we employ NeuralCoref\footnote{\url{https://github.com/huggingface/neuralcoref}} to detect the coreferent of the target word.
To address \textit{determiner-noun agreement}, we generate the dependency tree for each sample and extract the determiners corresponding to the target noun. Lastly, for \textit{subject-verb agreement}, the same dependency tree can be used to identify the subjects associated with the verb.

\paragraph{Model.}
In this section, we employ the RoBERTa \citep{liu2019robertarobustlyoptimizedbert} model for our evaluations.
We use both pre-trained\footnote{The results of the pre-trained model are covered in Section~\ref{sec:experiment2_apx}.} and fine-tuned versions of the model.
For fine-tuning, the target token is replaced with \texttt{[MASK]}, and the model is optimized to select the correct target token (the grammatically appropriate word) over the foil token (a similar but grammatically incorrect alternative).
Next, for inference, we use the head-on-top, also known as the unembedding matrix, to generate probabilities over the vocabulary.

\paragraph{Attribution Methods.} 
GlobEnc, ALTI, and Value Zeroing are the attribution methods we consider for comparison in this experiment.
Unlike Value Zeroing, which produces layer-wise attributions, the other two methods generate global importance scores.
To acquire per-layer attributions for GlobEnc and ALTI, we bypass the rollout aggregation method to directly derive per-layer scores and we denote them as GlobEnc¬ and ALTI¬.
We also consider a random baseline in which tokens are attributed equal scores.

\paragraph{Layer-wise Attribution with NormXLogit.}
Following the procedure described in Subsection~\ref{sec:LogAt}, we treat the language modeling setup as a classification task, where the number of labels corresponds to the size of the vocabulary.
In this setup, applying the head-on-top to the token representations yields a probability distribution over the vocabulary for each input token.
For NormXLogit, to obtain the attributions of layer $l$, we apply LogAt to the output representations of the $l$-th layer, combined with the $\ell^2$ norm of the input embeddings, where $l \in \{1, 2, \dots, L\}$.
Then, for each input token, $LogAt(target)$ is considered its attribution score with respect to the target token.
Thanks to the per-label attributions obtained via LogAt, we can also identify the importance of evidence words in predicting both foil and target tokens.
These attributions can be generated for any token in the vocabulary (cf. \ref{sec:experiment2_apx}).

\paragraph{Alignment Metrics.}
Following \citet{yin2022interpreting}, we define the known evidence (i.e., ground truth token-level rationale) as a binary vector $\bm{\mathcal{E}}$ with a length equal to the input sequence $X$.
In this vector, a value of $1$ at a given index indicates the presence of known evidence, while a value of $0$ indicates its absence.
Similarly, the explanation vector $\bm{\mathcal{S}}$ has the same length, with each element $\mathcal{S}_i$ representing the attribution score assigned to the $i$-th token for predicting the target token.
To evaluate the alignment between evidence and explanation vectors, we take advantage of the Dot Product and Average Precision metrics (see Section~\ref{sec:alignment_example} for a worked example).

\paragraph{Dot Product:} 
The dot product $\bm{\mathcal{E}} \cdot \bm{\mathcal{S}}$ measures the total score that the target attribution method assigns to the evidence tokens.

\paragraph{Average Precision:} 
To evaluate whether an attribution method successfully identifies the most important tokens, we use Average Precision (AP). This metric concentrates on the ranking obtained via the attribution method rather than its raw scores. For a given sample, AP is computed as:
\begin{equation} \label{eq4}
\text{AP} = \sum_{k = 1}^{n} (R_k - R_{k-1}) P_k
\end{equation}
where $R_k$ and $P_k$ indicate the recall and precision at threshold $k$, and $n$ is the length of the input sequence.

\subsection{Results}
\begin{figure}[t]
    \centering
    \includegraphics[width= \linewidth]{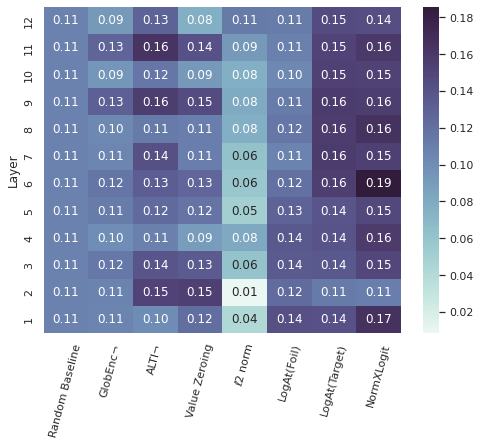}
    \caption{Per-layer alignment between evidence and explanation vectors in fine-tuned RoBERTa, calculated using Dot Product (higher values are better). Alignment for $\ell^2$ norm of word embeddings (layer 0): $0.14$.}
    \label{RoBERTa - fine-tuned Dot Product}
\end{figure}

\begin{figure}[t]
    \centering
    \includegraphics[width= \linewidth]{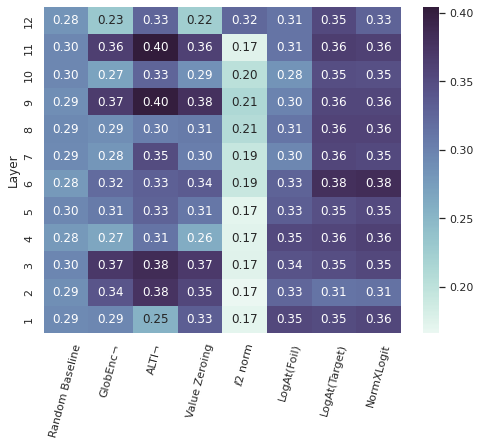}
    \caption{Per-layer alignment between evidence and explanation vectors in fine-tuned RoBERTa, calculated using Average Precision (higher values are better). Alignment for $\ell^2$ norm of word embeddings (layer 0): $0.35$.}
    \label{RoBERTa - fine-tuned Average Precision}
\end{figure}

Figures \ref{RoBERTa - fine-tuned Dot Product} and \ref{RoBERTa - fine-tuned Average Precision} present the results of the alignment for different attribution methods and the known evidence enforcing a linguistic paradigm.
In Figure \ref{RoBERTa - fine-tuned Dot Product}, it can be seen that across almost all layers, NormXLogit consistently outperforms other methods in the experiment.
The LogAt scores corresponding to the foil token in both alignment metrics are lower than those obtained from the target token.
Specifically, as we progress to higher Transformer layers, there is a drop in alignment for the foil token and an increase for the target token.
This pattern can be explained by the fact that token representations become more contextualized as they pass through layers.
Increased context mixing from evidence words can lead to a correct prediction ($LogAt(Target)$), while reduced context mixing can result in incorrect predictions ($LogAt(Foil)$).

As noted earlier, the LogAt scores can be calculated for other tokens in the vocabulary as well.
Our analysis shows that the LogAt score for the word `plural' ($LogAt(``plural")$) outperforms all other methods in our experiments by a notable margin.
This superior performance, unlike that of other random words, might be attributed to the \textit{number agreement} phenomena underlying this experiment.
At layer 7, the results of $LogAt(``plural")$ for Dot Product and Average Precision are $0.28$ and $0.50$, respectively (cf. \ref{sec:experiment2_apx}).

As mentioned in the caption of Figures \ref{RoBERTa - fine-tuned Dot Product} and \ref{RoBERTa - fine-tuned Average Precision}, the high alignment between the norm of input word embeddings and the evidence confirms that indeed they are informative.

\section{Conclusion and Future Work}
In this paper, we introduced NormXLogit, an architecture-agnostic interpretation method that can be applied to any setup to reveal the opaque mechanism behind the decision-making process of LLMs.
This method is fast and scalable, and it can be applied to models of any size.
By utilizing the head-on-top, we gain the advantage of producing per-label explanations, which can be used to identify the most important tokens with respect to each label.
Through extensive experiments, we showed that the attributions produced by NormXLogit are not only more faithful than those generated by gradient-based methods but also competitive with architecture-specific approaches.

Future work could explore the applicability of our proposed method to other domains and models, such as vision and non-Transformer architectures.
Another promising direction is to investigate how aggregating attributions across all labels in a classification setup could lead to improved explanations.

\section*{Limitations}
As with most attempts to interpret deep learning models, our evaluation, as well as those of existing methods, is inherently constrained by the lack of a definitive gold standard for understanding model internals. To mitigate this, we employed independent and complementary evaluation frameworks, including the AOPC metric, to measure the faithfulness of our approach.

Our work primarily targets models with a transformer architecture, with evaluations focused on models applied to text. Nevertheless, we believe our method is broadly applicable to scenarios where inputs are represented numerically and the model generates token-wise representations.

\bibliography{custom}

\clearpage

\appendix

\section{Appendix}

\subsection{Computing Infrastructure}
All experiments were conducted on a machine with an Nvidia Quadro RTX 8000 GPU with 48GB of memory.
The operating system used is Ubuntu 22.04.3 LTS.

\subsection{Experiment 1: Complete Results}
\label{sec:experiment1_apx}
\begin{figure}[h]
    \centering
    \includegraphics[width= \linewidth]{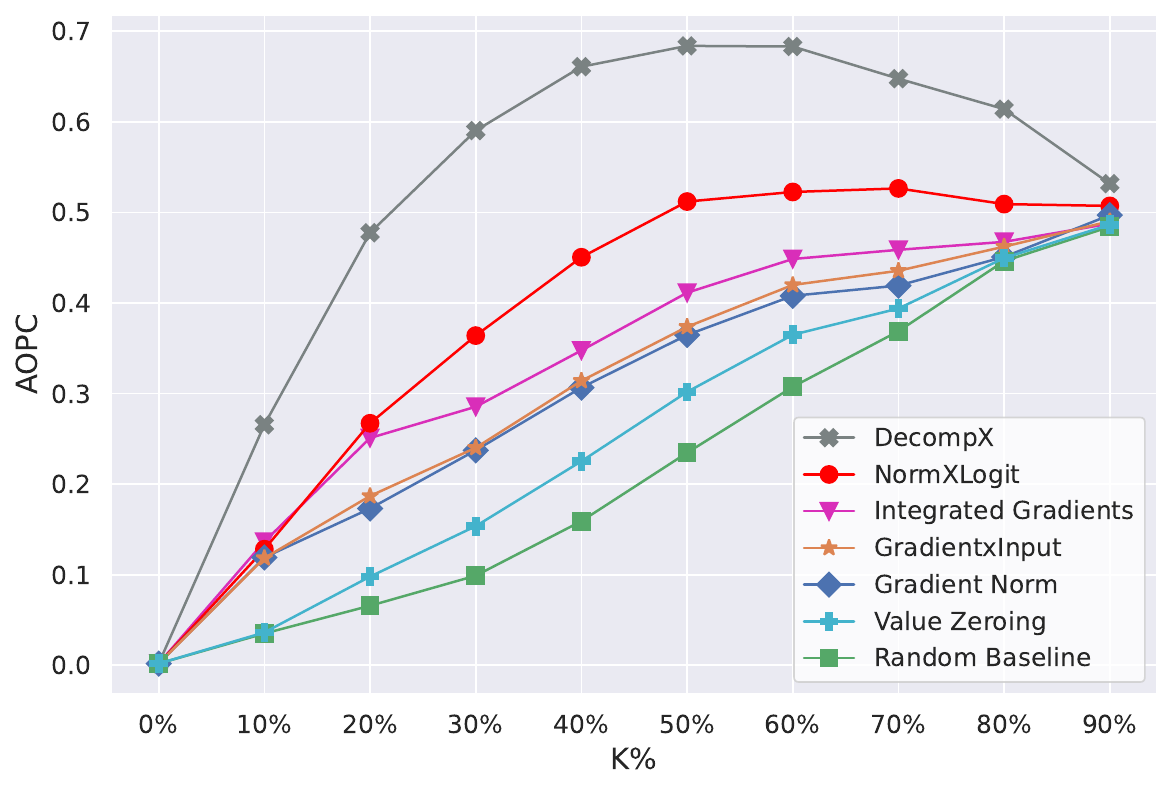}
    \caption{AOPC of different attribution methods for BERT fine-tuned on SST-2 (higher AOPC is better).}
    \label{SST2-BERT-norm2-AOPC}
\end{figure}

\begin{figure}[h]
    \centering
    \includegraphics[width= \linewidth]{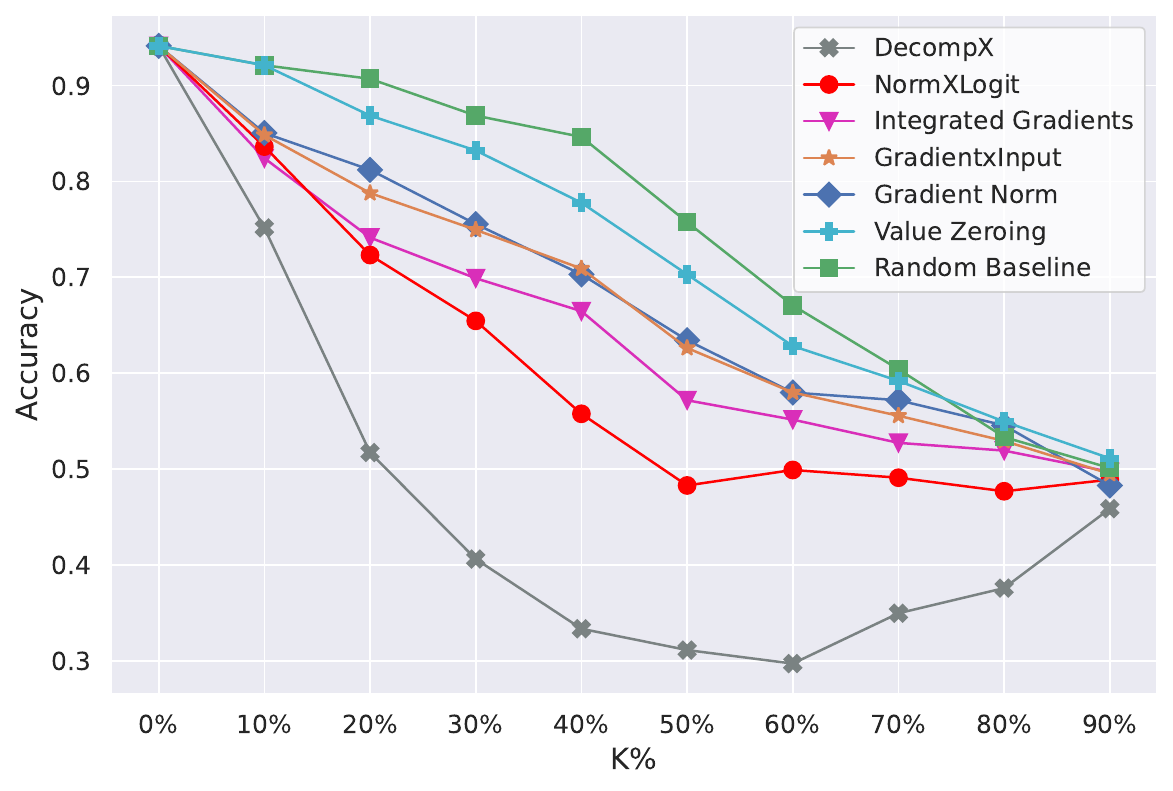}
    \caption{Accuracy of different attribution methods for BERT fine-tuned on SST-2 (lower Accuracy is better).}
    \label{SST2-BERT-norm2-Acc}
\end{figure}

\begin{figure}[h!]
    \centering
    \includegraphics[width= \linewidth]{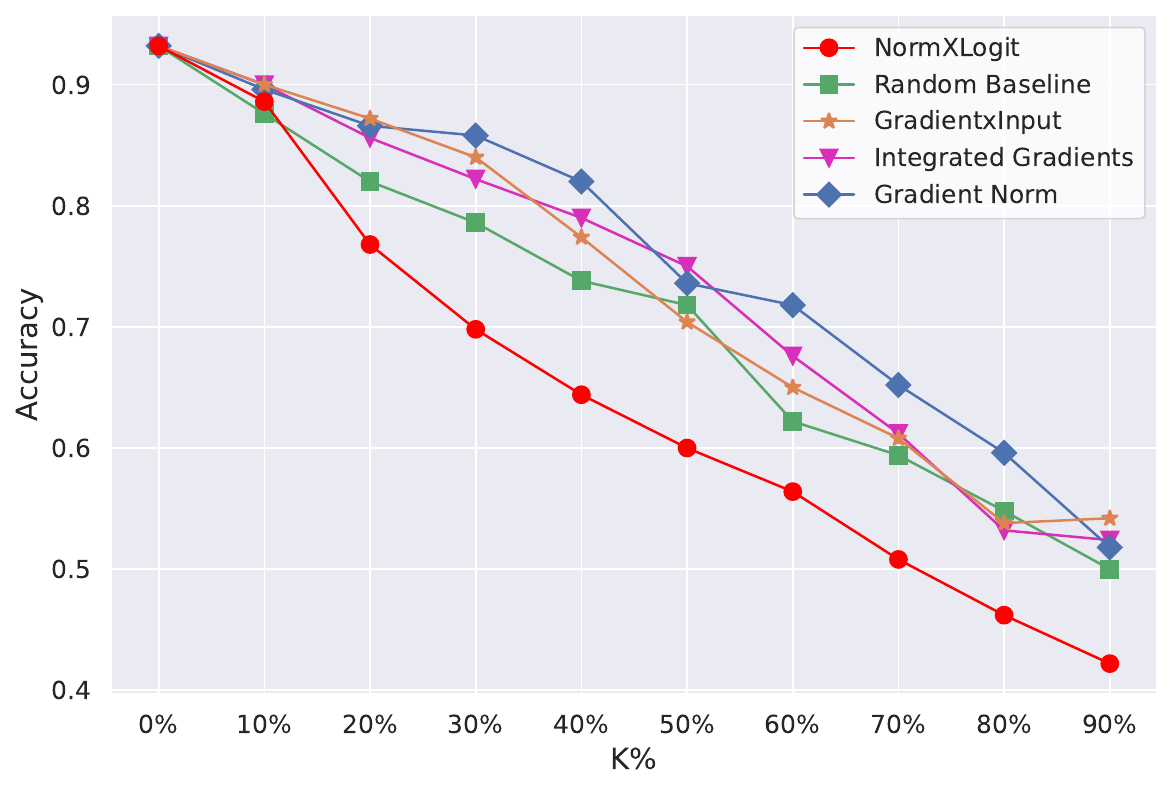}
    \caption{Accuracy of different attribution methods for \textsc{Llama2} fine-tuned on SST-2 (lower Accuracy is better).}
    \label{SST2-Llama2-norm2-Acc}
\end{figure}

\begin{figure}[h]
    \centering
    \includegraphics[width= \linewidth]{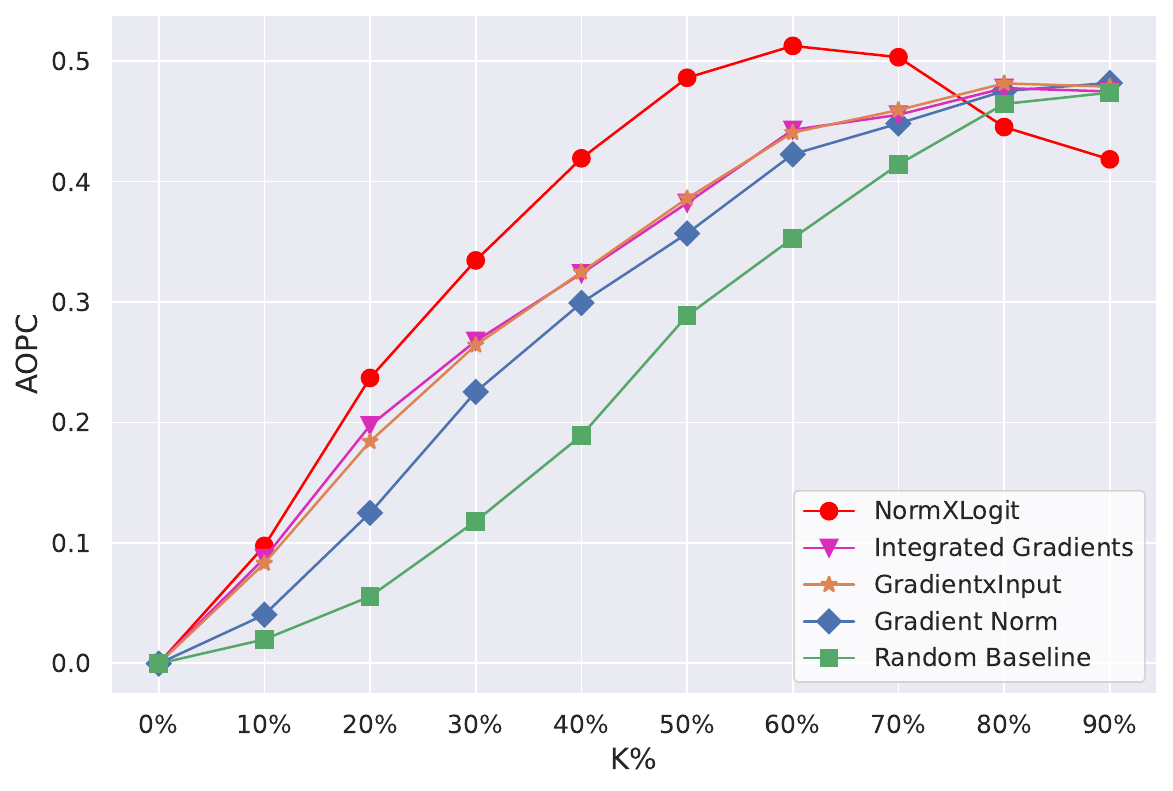}
    \caption{AOPC of different attribution methods for DeBERTa fine-tuned on SST-2 (higher AOPC is better).}
    \label{SST2-DeBERTa-norm2-AOPC}
\end{figure}

\begin{figure}[h!]
    \centering
    \includegraphics[width= \linewidth]{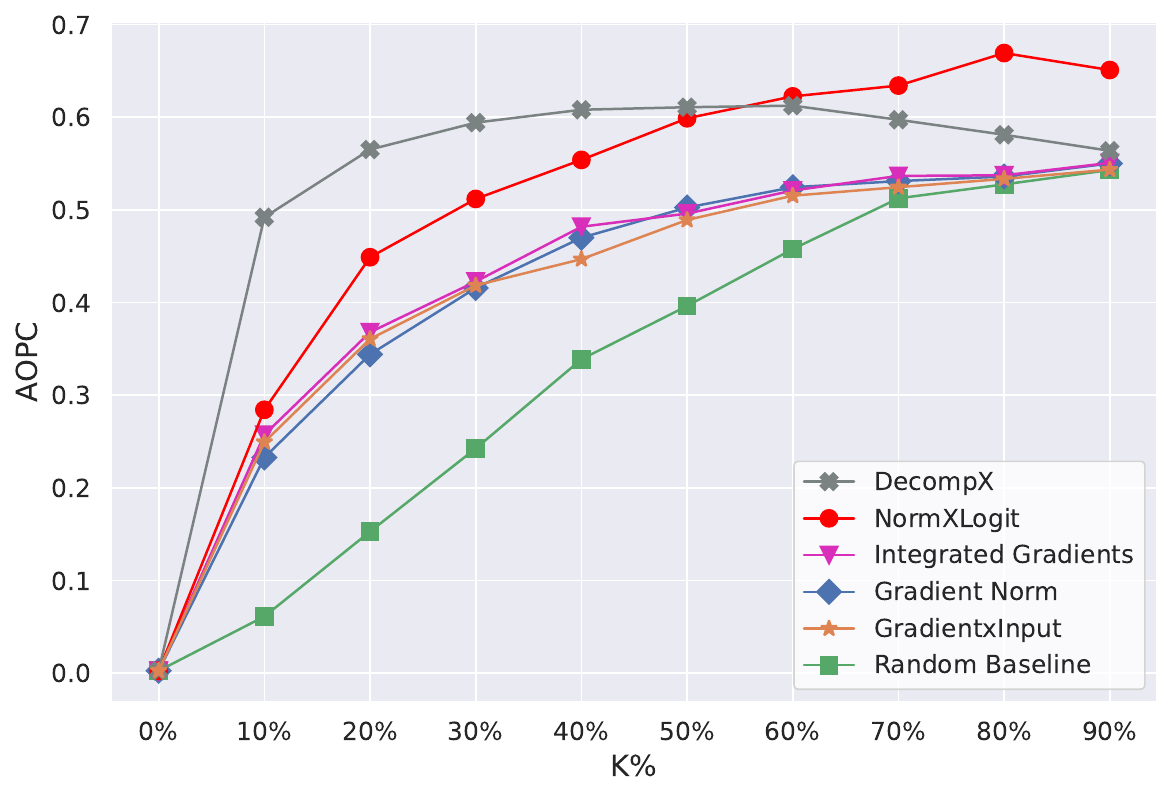}
    \caption{AOPC of different attribution methods for BERT fine-tuned on MultiNLI (higher AOPC is better).}
    \label{MNLI-BERT-Norm2-AOPC}
\end{figure}

\begin{figure}[h!]
    \centering
    \includegraphics[width= \linewidth]{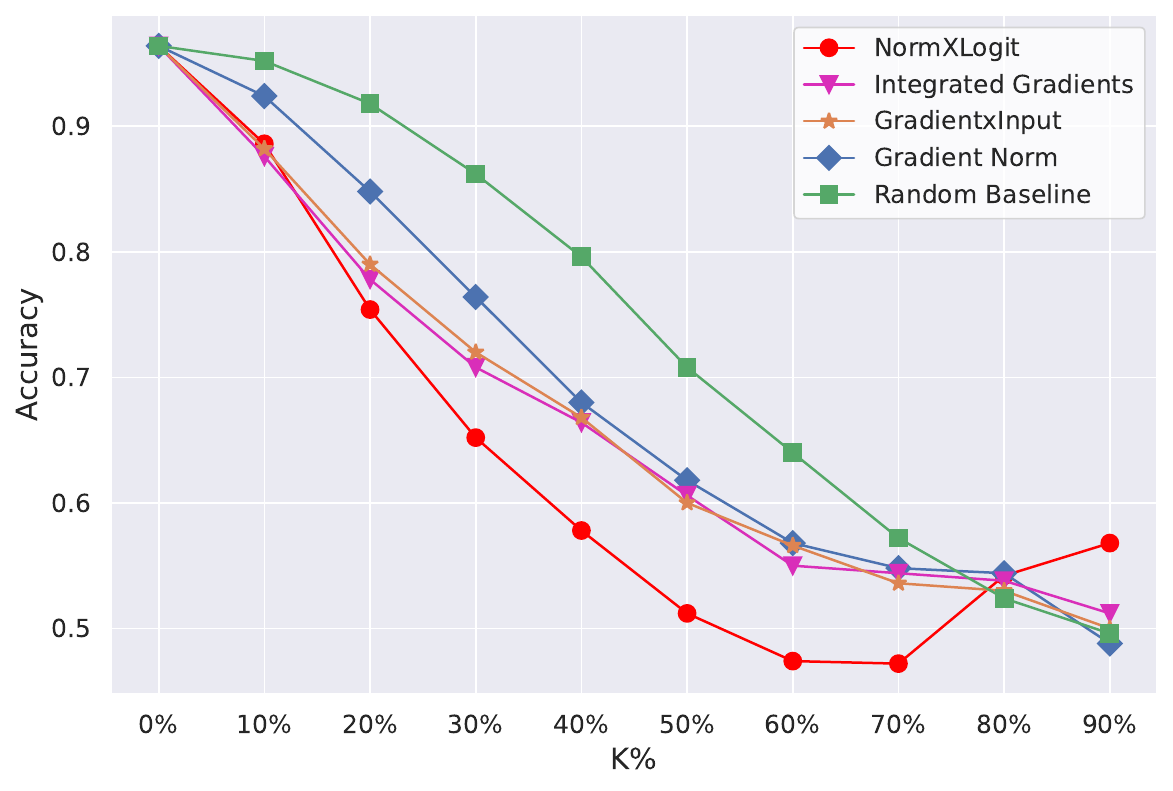}
    \caption{Accuracy of different attribution methods for DeBERTa fine-tuned on SST-2 (lower Accuracy is better).}
    \label{SST2-DeBERTa-norm2-Acc}
\end{figure}

\begin{figure}[h!]
    \centering
    \includegraphics[width= \linewidth]{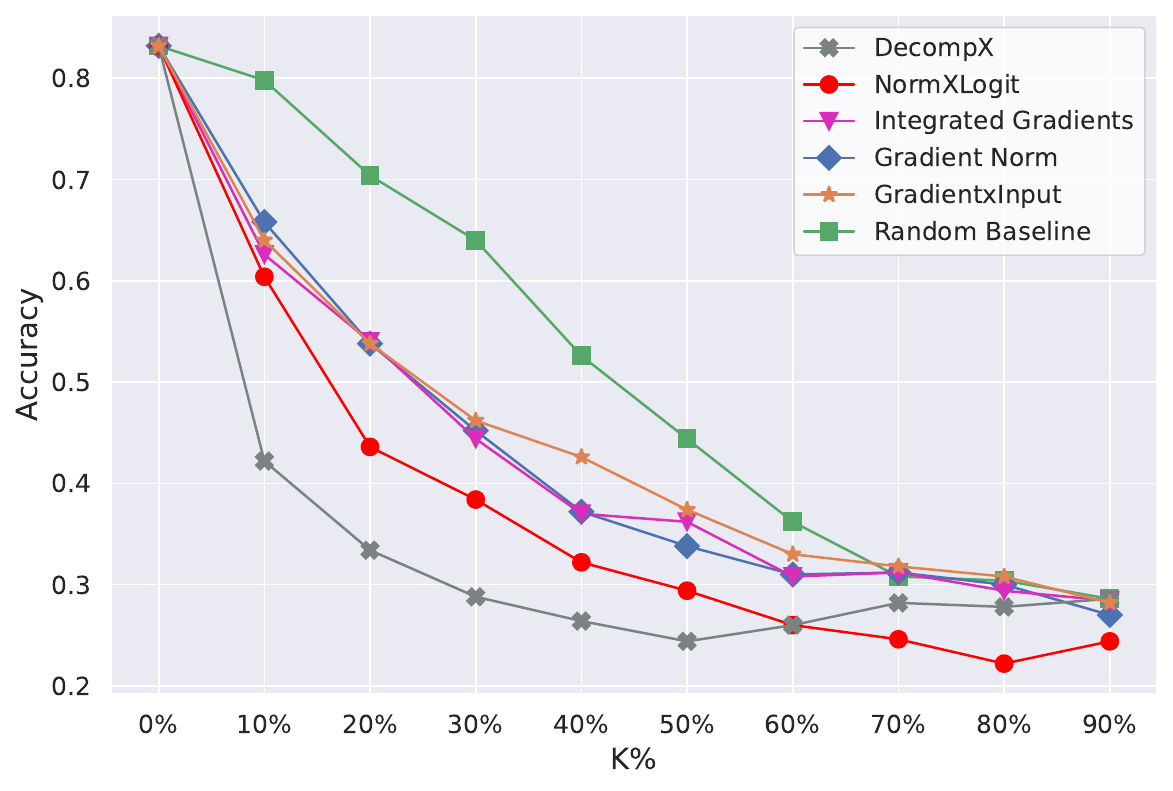}
    \caption{Accuracy of different attribution methods for BERT fine-tuned on MultiNLI (lower Accuracy is better).}
    \label{MNLI-BERT-Norm2-Acc}
\end{figure}

\begin{figure}[h!]
    \centering
    \includegraphics[width= \linewidth]{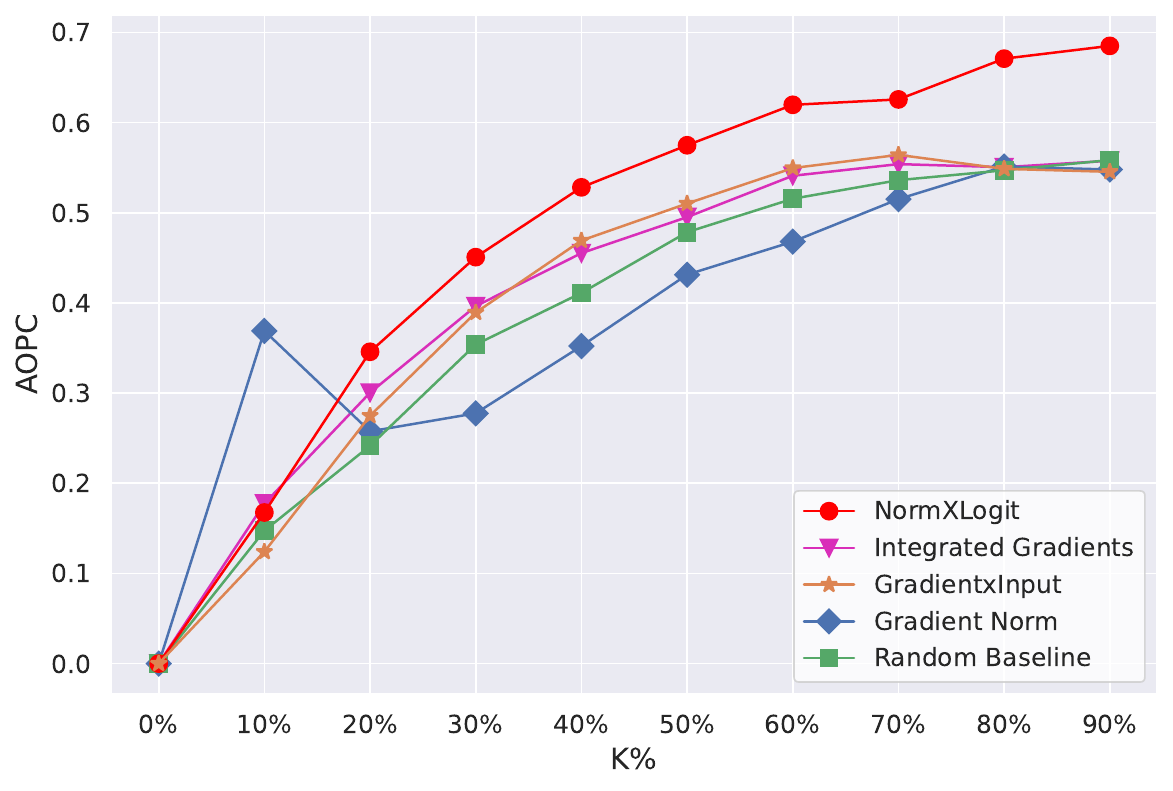}
    \caption{AOPC of different attribution methods for \textsc{Llama2} fine-tuned on MultiNLI (higher AOPC is better).}
    \label{MNLI-Llama-Norm2-AOPC}
\end{figure}

\begin{figure}[h!]
    \centering
    \includegraphics[width= \linewidth]{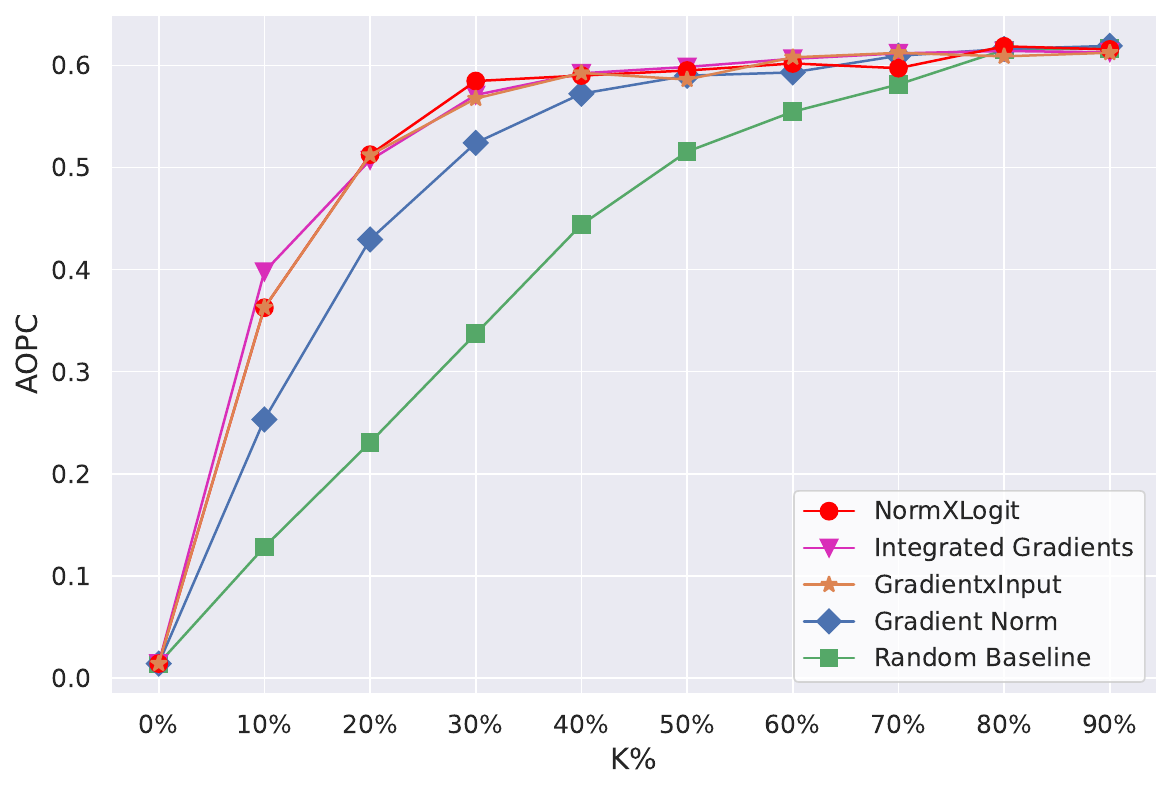}
    \caption{AOPC of different attribution methods for DeBERTa fine-tuned on MultiNLI (higher AOPC is better).}
    \label{MNLI-DeBERTa-Norm2-AOPC}
\end{figure}

\begin{figure}[h!]
    \centering
    \includegraphics[width= \linewidth]{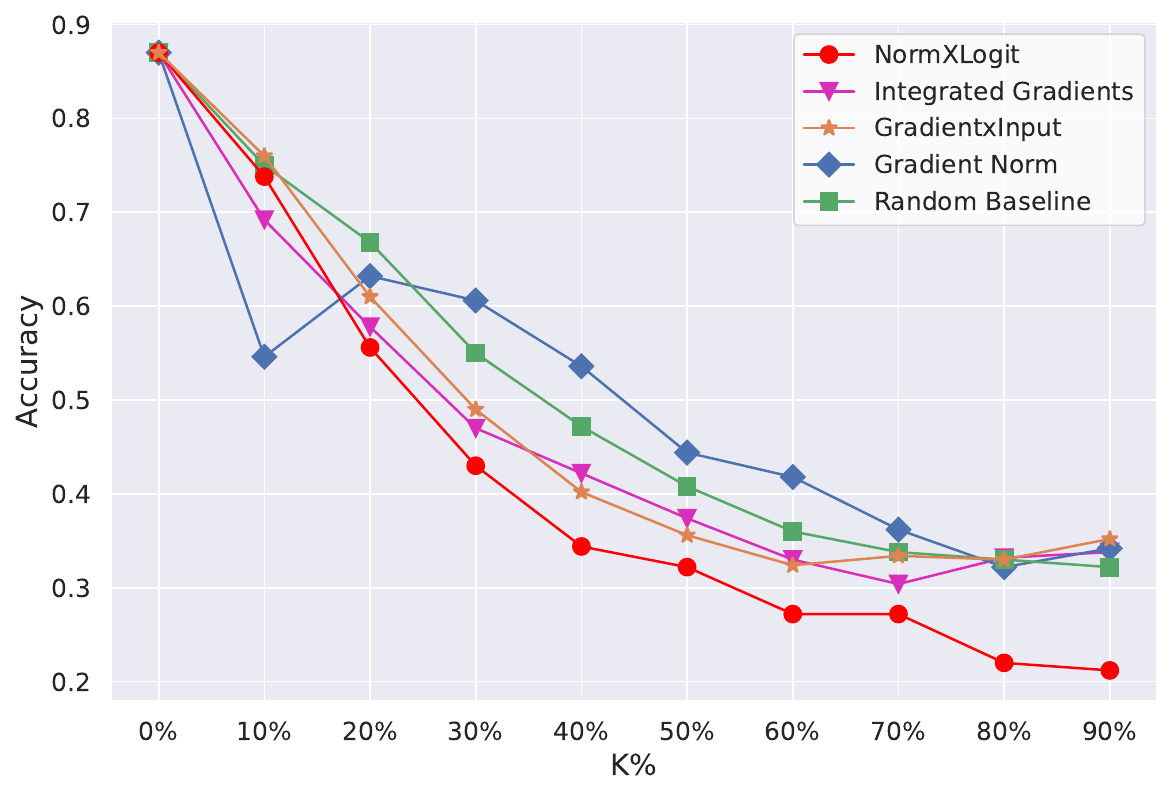}
    \caption{Accuracy of different attribution methods for \textsc{Llama2} fine-tuned on MultiNLI (lower Accuracy is better).}
    \label{MNLI-Llama-Norm2-Acc}
\end{figure}

\begin{figure}[h!]
    \centering
    \includegraphics[width= \linewidth]{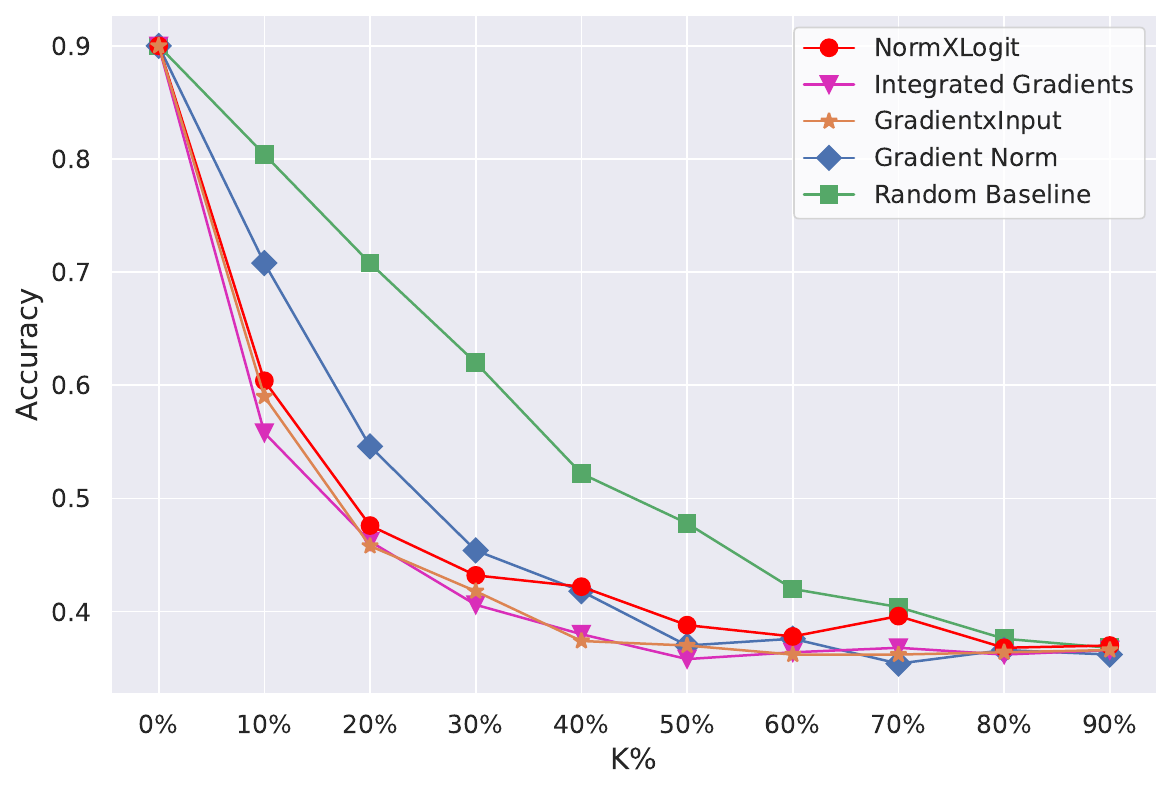}
    \caption{Accuracy of different attribution methods for DeBERTa fine-tuned on MultiNLI (lower Accuracy is better).}
    \label{MNLI-DeBERTa-Norm2-Acc}
\end{figure}

\begin{figure}[h!]
    \centering
    \includegraphics[width= \linewidth]{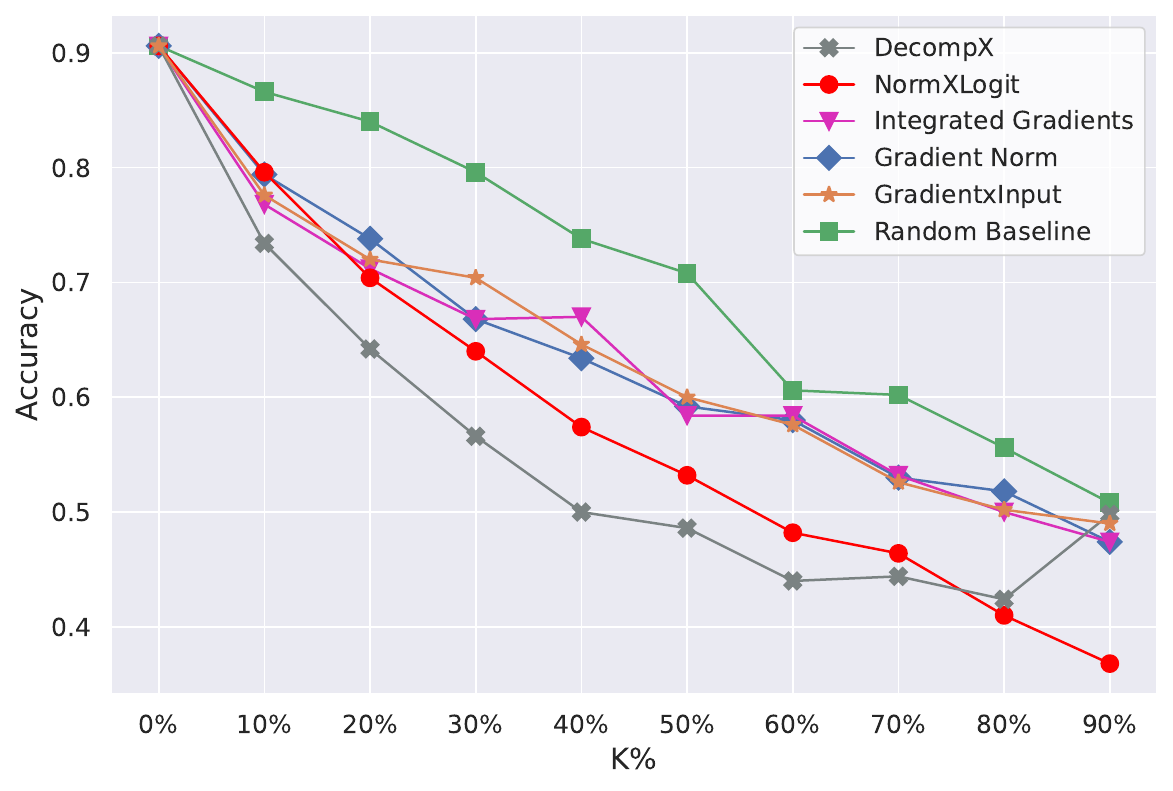}
    \caption{Accuracy of different attribution methods for BERT fine-tuned on QNLI (lower Accuracy is better).}
    \label{QNLI-BERT-Norm2-Acc}
\end{figure}

\begin{figure}[h!]
    \centering
    \includegraphics[width= \linewidth]{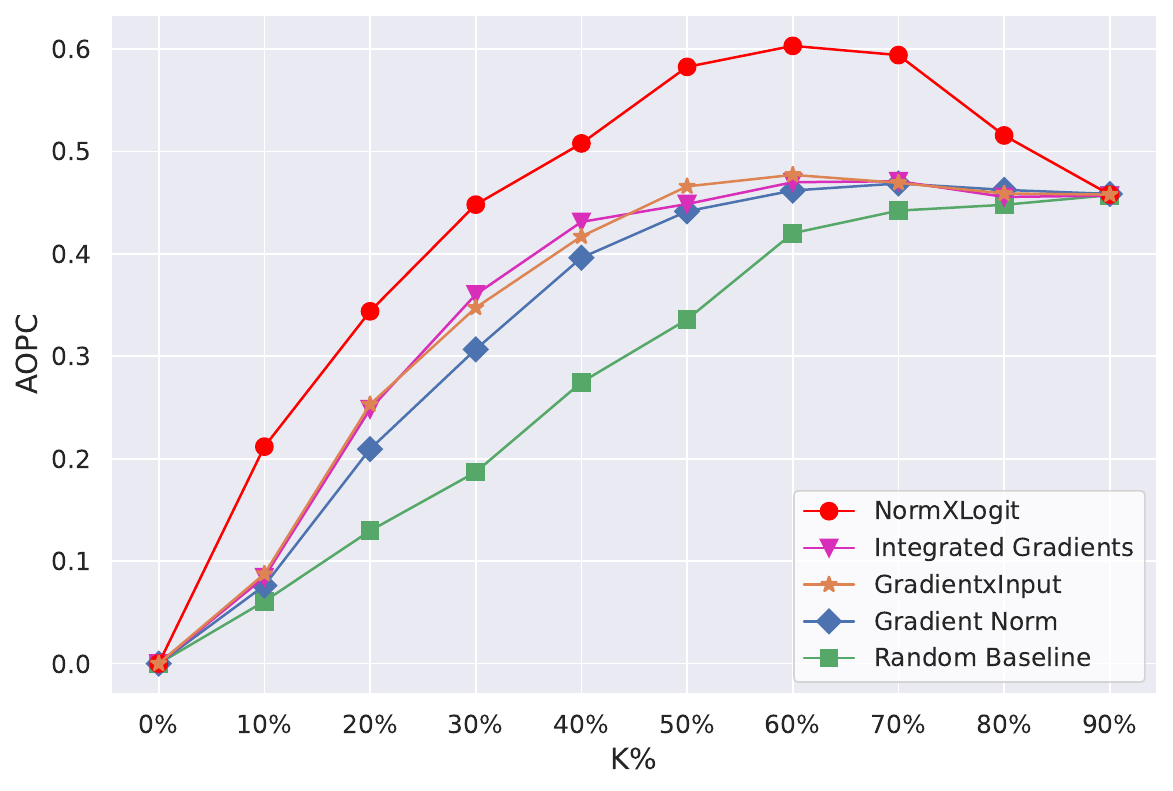}
    \caption{AOPC of different attribution methods for DeBERTa fine-tuned on QNLI (higher AOPC is better).}
    \label{QNLI-DeBERTa-Norm2-AOPC}
\end{figure}

\begin{figure}[h!]
    \centering
    \includegraphics[width= \linewidth]{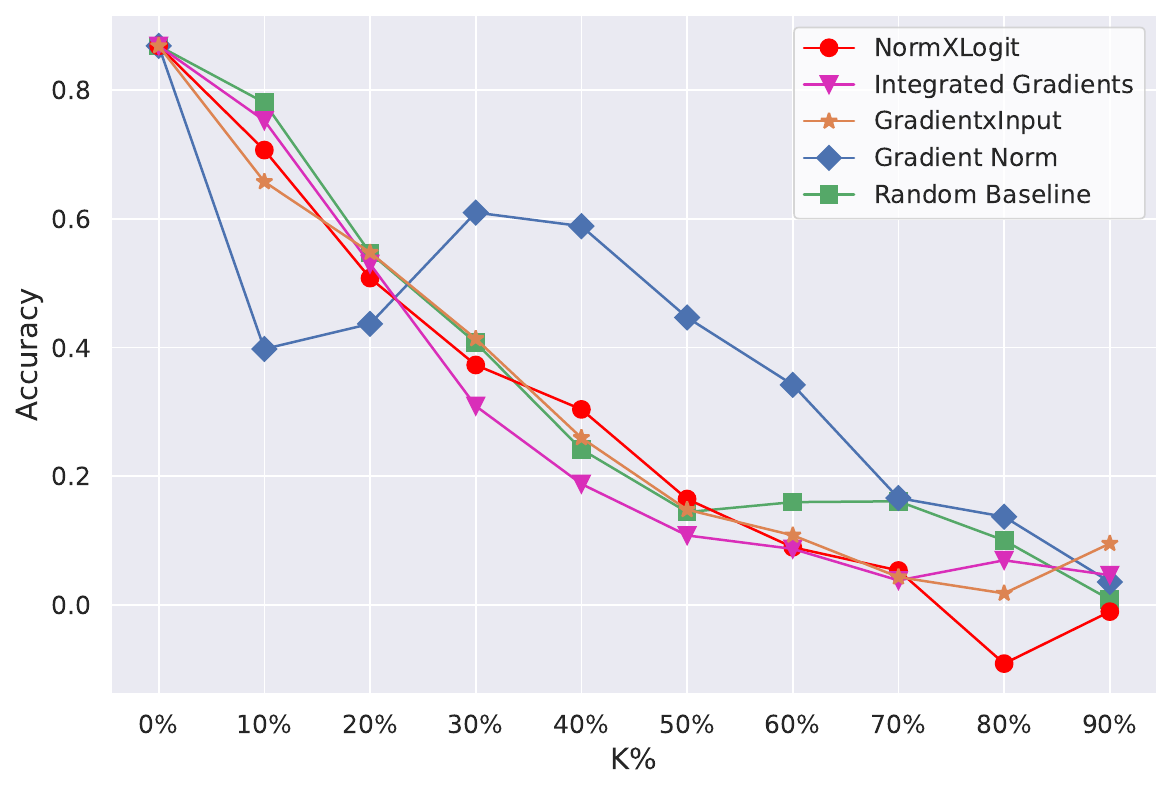}
    \caption{Accuracy of different attribution methods for \textsc{Llama2} fine-tuned on STS-B (lower Accuracy is better).}
    \label{STS-B-Llama-Norm2-Acc}
\end{figure}

\begin{figure}[h!]
    \centering
    \includegraphics[width= \linewidth]{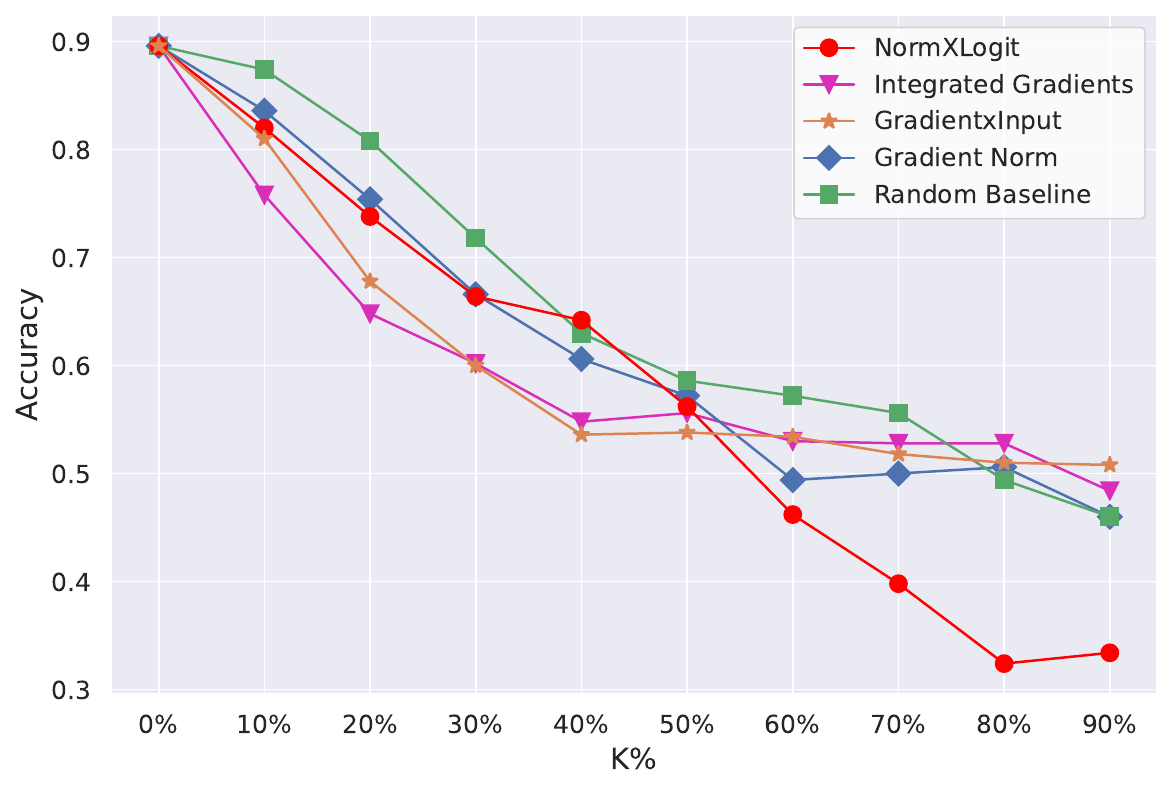}
    \caption{Accuracy of different attribution methods for \textsc{Llama2} fine-tuned on QNLI (lower Accuracy is better).}
    \label{QNLI-Llama-Norm2-Acc}
\end{figure}

\begin{figure}[h!]
    \centering
    \includegraphics[width= \linewidth]{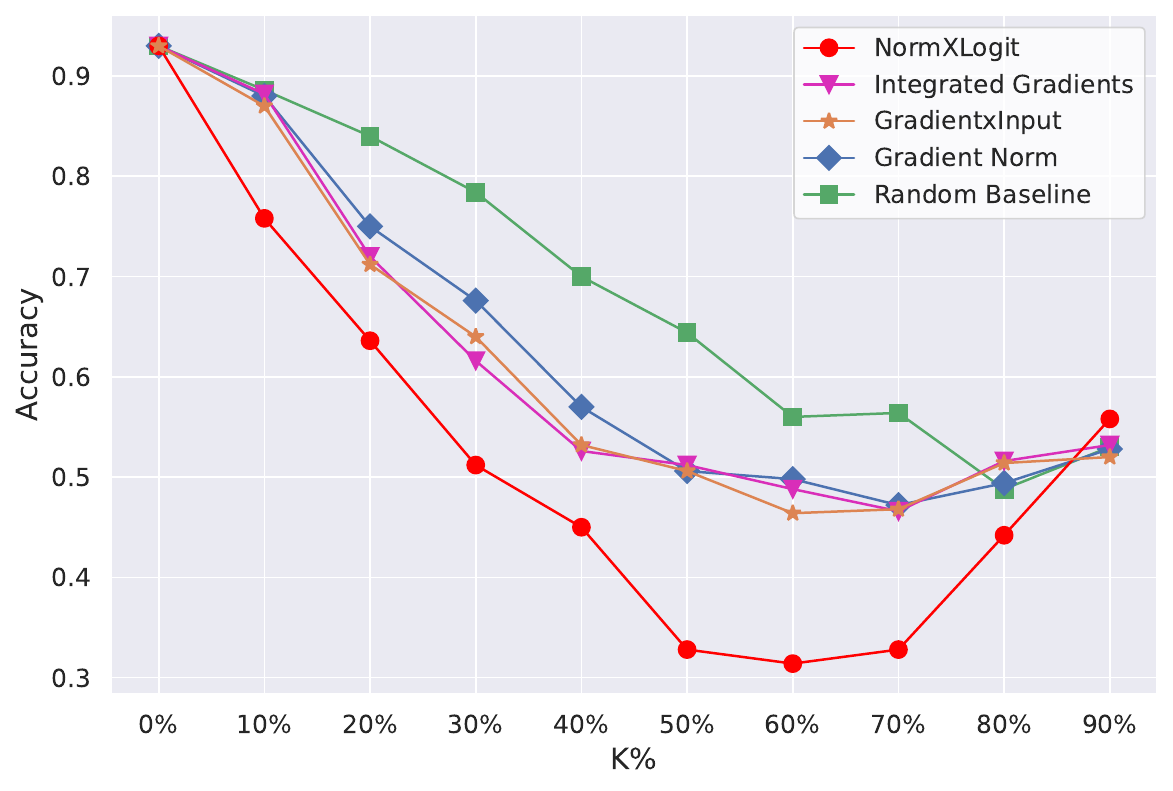}
    \caption{Accuracy of different attribution methods for DeBERTa fine-tuned on QNLI (lower Accuracy is better).}
    \label{QNLI-DeBERTa-Norm2-Acc}
\end{figure}

\begin{figure}[h!]
    \centering
    \includegraphics[width= \linewidth]{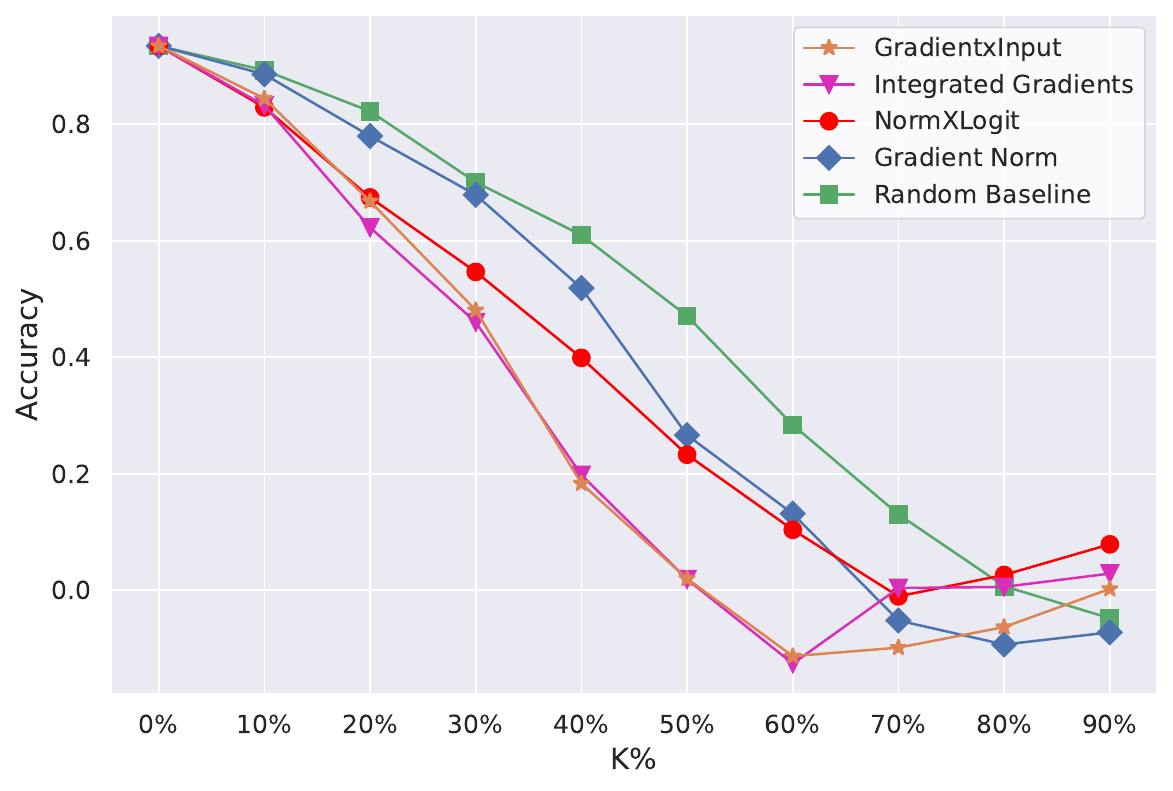}
    \caption{Accuracy of different attribution methods for DeBERTa fine-tuned on STS-B (lower Accuracy is better).}
    \label{STS-B-DeBERTa-Norm2-Acc}
\end{figure}

Figures \ref{SST2-BERT-norm2-AOPC} to \ref{STS-B-DeBERTa-Norm2-Acc} illustrate the AOPC and Accuracy across different models and datasets.
In Figure \ref{SST2-BERT-norm2-AOPC}, we demonstrate the global performance of Value Zeroing on the SST-2 dataset.
The results show that this method is not faithful to the model's decision-making process.
This issue may stem from the inherent limitations of the rollout aggregation method, as previously discussed.
Additionally, since Value Zeroing is a perturbation-based method, it may also inherit some of the challenges associated with these approaches.
For instance, this method zeros out each token's value vector one at a time, which can lead to problems like ignoring dependencies between features.
Consider the following example:

\begin{quote}
\textit{"The movie is mediocre, maybe even bad."}
\end{quote}
In this case, erasing “\textit{mediocre}" or “\textit{bad}" independently may not significantly impact the overall sentiment of the sentence.

For our Integrated Gradients experiments, we generally used 50 steps. However, for \textsc{Llama2}, we reduced the number of steps to 25 due to resource constraints.

\paragraph{Computational Efficiency.}
We conducted experiments to compare the computational efficiency of the attribution methods, using the BERT model with a maximum input length of 512 tokens. Timing results, averaged over five runs and reported in seconds, reveal that the runtime for Gradient Norm and Gradient$\times$Input remains nearly constant across different input lengths, while Integrated Gradients and DecompX show noticeable increases as input length grows. NormXLogit consistently demonstrates the fastest processing times, with minimal sensitivity to input length. In terms of memory efficiency, we identified the maximum batch size each method could handle within 48GB of GPU memory. NormXLogit significantly outperforms others, processing up to 750 samples per batch, whereas Integrated Gradients and DecompX are limited to very small batch sizes (2 and 1, respectively). Consequently, NormXLogit offers both superior speed and memory efficiency, enabling more scalable attribution computations.

\begin{figure}[h!]
    \centering
    \includegraphics[width= \linewidth]{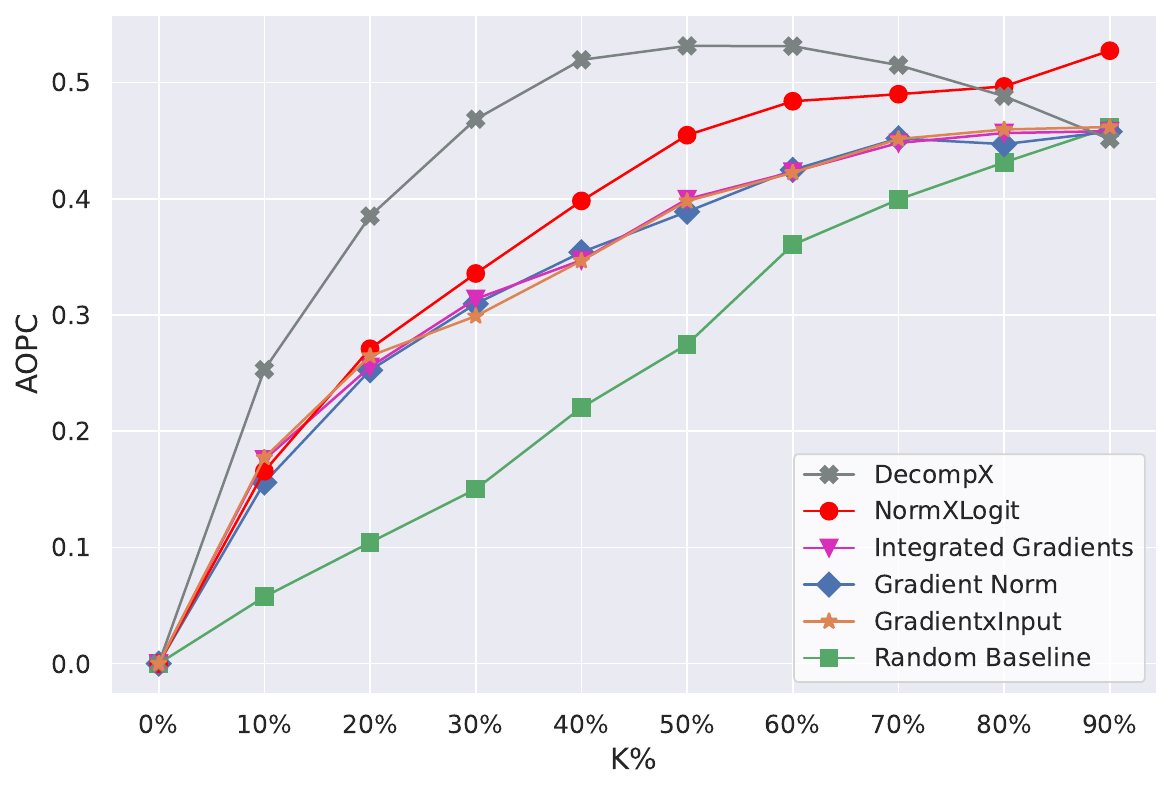}
    \caption{AOPC of different attribution methods for BERT fine-tuned on QNLI (higher AOPC is better).}
    \label{QNLI-BERT-Norm2-AOPC}
\end{figure}

\begin{figure}[h!]
    \centering
    \includegraphics[width= \linewidth]{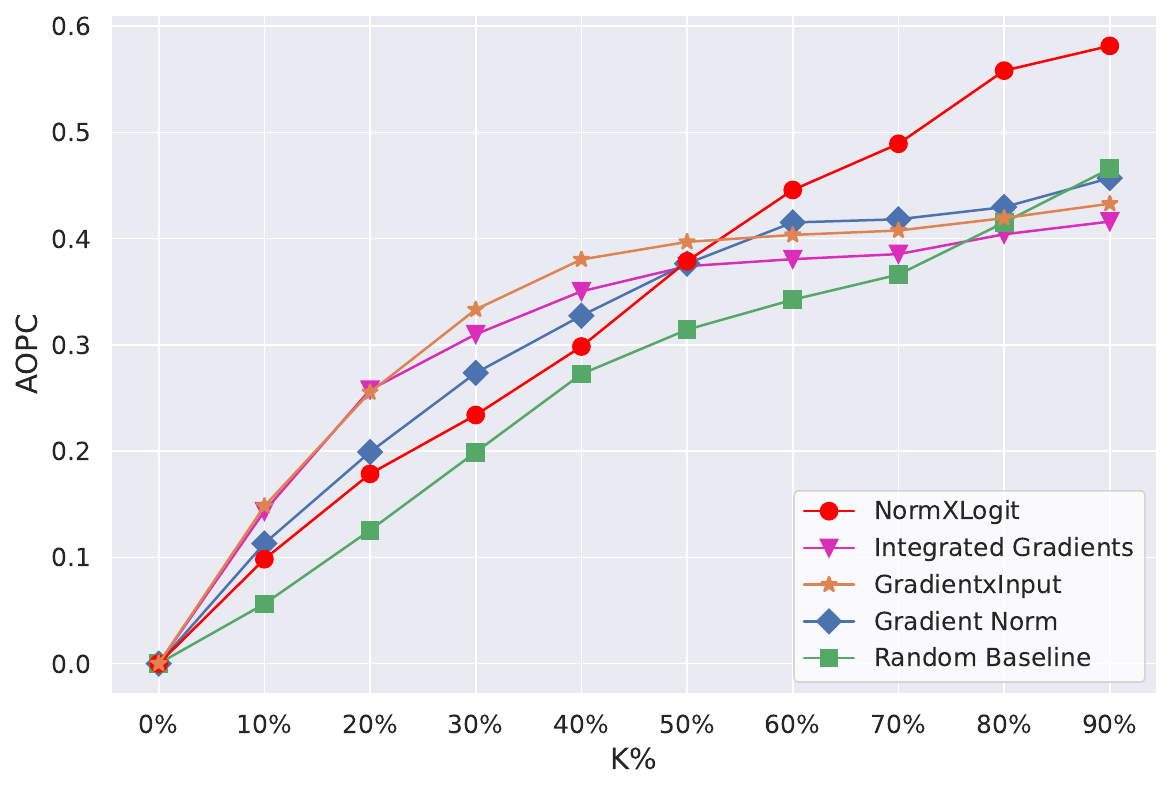}
    \caption{AOPC of different attribution methods for \textsc{Llama2} fine-tuned on QNLI (higher AOPC is better).}
    \label{QNLI-BERT-Norm2-AOPC}
\end{figure}

\begin{table*}[h]
\begin{center}
\small
\tabcolsep=0.13cm
\begin{tabular}{l c c c | c c c | c c c} 
 \toprule
     & 
     \multicolumn{3}{c}{\textbf{\textsc{SST-2}} \scriptsize{\text{\textsc{(Acc$\downarrow$)}}}} & 
     \multicolumn{3}{c}{\textbf{\textsc{MNLI}} \scriptsize{\text{\textsc{(Acc$\downarrow$)}}}} & 
     \multicolumn{3}{c}{\textbf{\textsc{QNLI}} \scriptsize{\text{\textsc{(Acc$\downarrow$)}}}} \\
     \cmidrule(lr){2-10}

    & \scriptsize{\textbf{\textsc{Llama 2}}} & \scriptsize{\textbf{DeBERTa}} & \scriptsize{\textbf{BERT}}
    & \scriptsize{\textbf{\textsc{Llama 2}}} & \scriptsize{\textbf{DeBERTa}} & \scriptsize{\textbf{BERT}}
    & \scriptsize{\textbf{\textsc{Llama 2}}} & \scriptsize{\textbf{DeBERTa}} & \scriptsize{\textbf{BERT}} \\
    \midrule
    Random Baseline
& 0.793 & 0.826 & 0.839 & 0.563 & 0.622 & 0.578 & 0.733 & 0.770 & 0.792 \\
    Gradient Norm
& 0.844 & 0.772 & 0.764 & 0.564 & 0.539 & 0.487 & 0.699 & 0.700 & 0.715 \\
    Gradient×Input
& 0.818 & 0.751 & 0.758 & 0.536 & 0.507 & 0.501 & 0.681 & 0.684 & 0.716 \\
    Integrated Gradients
& 0.822 & 0.749 & 0.726 & 0.523 & \textbf{0.503} & 0.486 & 0.675 & 0.688 & 0.711 \\
    DecompX
& N/A & N/A & \textbf{0.527} & N/A & N/A & \textbf{0.388} & N/A & N/A & \textbf{0.627} \\
    \midrule
    $\ell^2$ norm
& 0.743 & 0.731 & 0.775 & 0.592 & 0.602 & 0.572 & 0.748 & 0.736 & 0.768 \\
LogAt
& 0.768 & 0.791 & 0.757 & 0.554 & 0.599 & 0.545 & 0.715 & 0.724 & 0.745 \\
    \textbf{NormXLogit}
& \textbf{0.720} & \textbf{0.711} & 0.684 & \textbf{0.471} & 0.526 & 0.427 & \textbf{0.649} & \textbf{0.584} & 0.653 \\
 \bottomrule
\end{tabular}
\end{center}
\caption{
Performance evaluation of NormXLogit against other methods across various model and dataset configurations.
Each value is computed by averaging across all perturbation ratios (lower Accuracy is better).
}
\label{Tab:Acc}
\end{table*}

\begin{table*}[h]
\begin{center}
\small
\tabcolsep=0.13cm
\begin{tabular}{l c c c c | c}
 \toprule
     & 
     \multicolumn{5}{c}     {\textbf{\textsc{Method}}} \\
     \cmidrule(lr){2-6}

    \textbf{Input Length}
    & \makebox[1.8cm][c]{\scriptsize{\textbf{Gradient Norm}}} & \makebox[1.8cm][c]{\scriptsize{\textbf{Gradient×Input}}} & \makebox[2cm][c]{\scriptsize{\textbf{Integrated Gradients}}}
     & \makebox[1.8cm][c]{\scriptsize{\textbf{DecompX}}} & \makebox[1.8cm][c]{\scriptsize{\textbf{NormXLogit}}} \\
    \midrule
    40
& 0.77$_{\pm\text{0.03}}$ & 0.80$_{\pm\text{0.02}}$ & 0.82$_{\pm\text{0.02}}$ & 0.39$_{\pm\text{0.01}}$ & \textbf{0.36}$_{\pm\text{0.00}}$ \\
    120
& 0.77$_{\pm\text{0.02}}$ & 0.81$_{\pm\text{0.02}}$ & 0.96$_{\pm\text{0.03}}$ & 0.38$_{\pm\text{0.00}}$ & \textbf{0.36}$_{\pm\text{0.00}}$ \\
    360
& 0.76$_{\pm\text{0.03}}$ & 0.80$_{\pm\text{0.02}}$ & 1.47$_{\pm\text{0.05}}$ & 0.99$_{\pm\text{0.02}}$ & \textbf{0.38}$_{\pm\text{0.01}}$ \\
    512
& 0.78$_{\pm\text{0.02}}$ & 0.80$_{\pm\text{0.02}}$ & 1.79$_{\pm\text{0.06}}$ & 2.36$_{\pm\text{0.02}}$ & \textbf{0.36}$_{\pm\text{0.00}}$ \\
 \bottomrule
\end{tabular}
\end{center}
\caption{
Average time (in seconds) per instance for different attribution methods across various input lengths.
NormXLogit demonstrates significantly lower computational costs compared to other methods and remains unaffected by input sequence length.
Results are averaged over five runs on BERT, and the values in subscript represent the standard deviation.
}
\label{Tab:time}
\end{table*}

\begin{table*}[h]
\begin{center}
\small
\tabcolsep=0.13cm
\begin{tabular}{l c c c c | c}
 \toprule
     & 
     \multicolumn{5}{c}     {\textbf{\textsc{Method}}} \\
     \cmidrule(lr){2-6}

    \textbf{Evaluation Criteria}
    & \makebox[1.8cm][c]{\scriptsize{\textbf{Gradient Norm}}} & \makebox[1.8cm][c]{\scriptsize{\textbf{Gradient×Input}}} & \makebox[2cm][c]{\scriptsize{\textbf{Integrated Gradients}}}
     & \makebox[1.8cm][c]{\scriptsize{\textbf{DecompX}}} & \makebox[1.8cm][c]{\scriptsize{\textbf{NormXLogit}}} \\
    \midrule
    Effective Batch Size
& 100 & 100 & 2 & 1 & \textbf{750} \\
    Average Time per Instance (s)
& 0.0076$_{\pm\text{0.00}}$ & 0.0081$_{\pm\text{0.00}}$ & 1.4104$_{\pm\text{0.01}}$ & 2.3625$_{\pm\text{0.02}}$ & \textbf{0.0005}$_{\pm\text{0.00}}$ \\
 \bottomrule
\end{tabular}
\end{center}
\caption{
Efficiency of attribution methods when maximizing batch size under a 48GB memory constraint (input length = 512).
NormXLogit supports significantly larger batch sizes and achieves the lowest per-instance time, demonstrating superior scalability and memory efficiency. Results are averaged over five runs on BERT, and the values in subscript represent the standard deviation.
}
\label{Tab:memory}
\end{table*}

\subsection{Experiment 2: RoBERTa Complete Results}
\label{sec:experiment2_apx}
\begin{figure}[h]
    \centering
    \includegraphics[width=\linewidth]{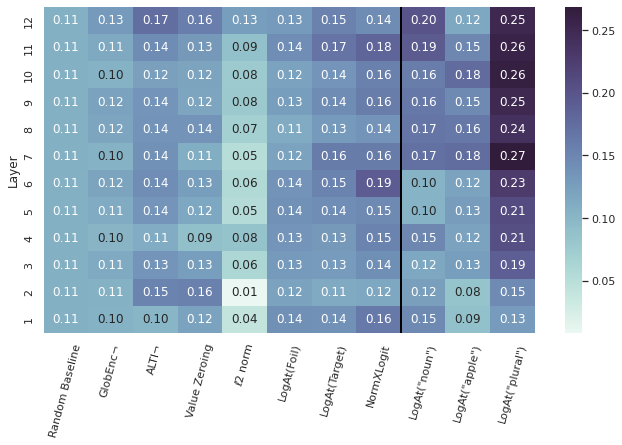}%
    \caption{Per-layer alignment between evidence and explanation vectors for the pre-trained version of RoBERTa, calculated using Dot Product metric (higher values are better). The alignment for $\ell^2$ norm of word embeddings (layer 0) is $0.14$.}
    \label{RoBERTa - pre-trained Dot Product (Extended)}
    \vspace{1cm}
    \includegraphics[width=\linewidth]{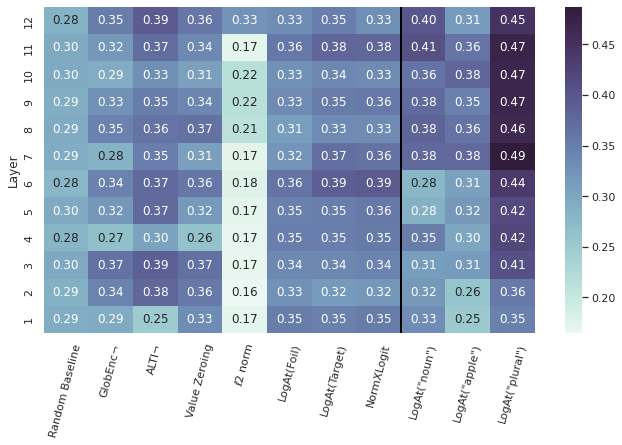}
    \caption{Per-layer alignment between evidence and explanation vectors for the pre-trained version of RoBERTa, calculated using Average Precision metric (higher values are better). The alignment for $\ell^2$ norm of word embeddings (layer 0) is $0.35$.}
    \label{RoBERTa - pre-trained Average Precision (Extended)}
\end{figure}

\begin{figure}[h]
    \centering
    \includegraphics[width=\linewidth]{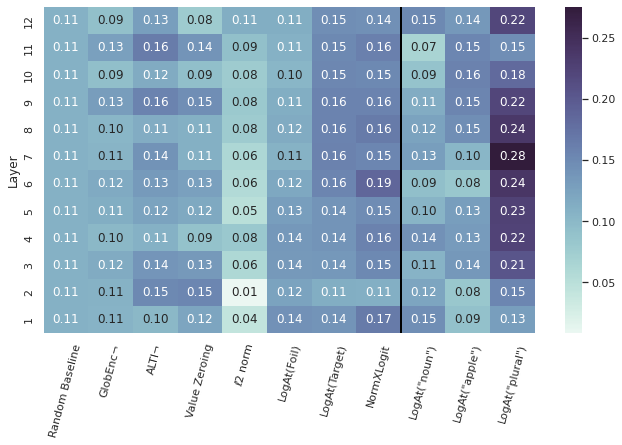}
    \caption{Per-layer alignment between evidence and explanation vectors for the fine-tuned version of RoBERTa, calculated using Dot Product metric (higher values are better). The alignment for $\ell^2$ norm of word embeddings (layer 0) is $0.14$.}
    \label{RoBERTa - fine-tuned Dot Product (Extended)}
    \vspace{1cm}
    \includegraphics[width=\linewidth]{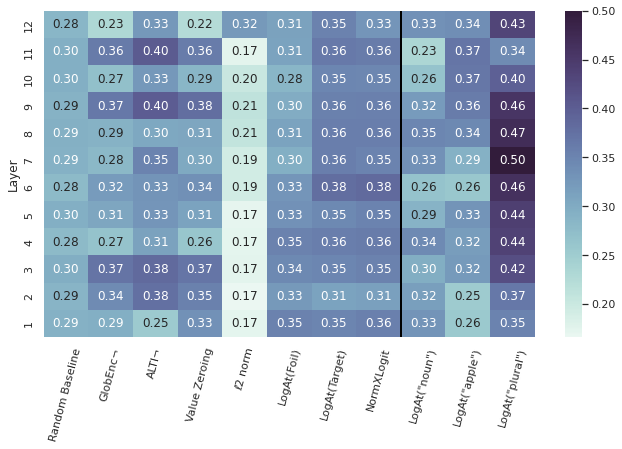}
    \caption{Per-layer alignment between evidence and explanation vectors for the fine-tuned version of RoBERTa, calculated using Average Precision metric (higher values are better). The alignment for $\ell^2$ norm of word embeddings (layer 0) is $0.35$.}
    \label{RoBERTa - fine-tuned Average Precision (Extended)}
    \label{fig:side-by-side}
\end{figure}

The evidence alignment experiment is conducted on a masked language modeling task to understand why a particular target token is chosen.
The LogAt method provides per-label attribution scores, enabling us to apply it to other labels (i.e., tokens in the vocabulary) to identify the most important tokens in the input sequence for predicting each specific label.
Figures \ref{RoBERTa - pre-trained Dot Product (Extended)} and \ref{RoBERTa - pre-trained Average Precision (Extended)} display the results of the Dot Product and Average Precision alignment metrics for the pre-trained RoBERTa model.
An important observation is the notable performance of $LogAt("plural")$, which demonstrates its effectiveness in identifying evidence words.
This level of performance is not seen with two other randomly chosen words.
Specifically, the results are more pronounced in the top layers, indicating that increased context mixing enhances the connection between the evidence and the word “plural."
In other words, as we progress through the layers, the contextualized representation of the evidence word becomes increasingly similar to the word “plural," resulting in a higher attribution for this word.
We attribute the superior performance for the word “plural" primarily to the nature of the phenomena used from the BLiMP dataset, which focused on \textit{number agreement}.
Figures \ref{RoBERTa - fine-tuned Dot Product (Extended)} and \ref{RoBERTa - fine-tuned Average Precision (Extended)} demonstrate the results for the fine-tuned RoBERTa model.

\clearpage

\clearpage

\subsection{Experiment 2: Worked Example for Alignment Metric Computation}
\label{sec:alignment_example}

To illustrate how the alignment metrics work, we provide an example using the following sentence:

\begin{quote}
\textit{"Karla thinks/think about it"}
\end{quote}

In this example, \textbf{``Karla''} serves as the evidence that determines the correct verb (\textit{``thinks''}).

\paragraph{Vector Definitions.}
We first define a binary evidence vector $\bm{\mathcal{E}}$ to indicate which tokens are considered known evidence (1 for evidence, 0 otherwise). For this example, the input sequence is tokenized as:

\[
\text{[Karla, thinks, about, it]}
\]

Since only ``Karla'' is the evidence, we define:
\[
\bm{\mathcal{E}} = [1, 0, 0, 0]
\]

Next, we define the attribution vector $\bm{\mathcal{S}}$, where each element $\mathcal{S}_i$ corresponds to the score assigned by the attribution method to token $i$:
\[
\bm{\mathcal{S}} = [0.3, 0.1, 0.5, 0.1]
\]

\paragraph{Dot Product:}
We compute the dot product between the two vectors:
\begin{align*}
\textstyle
\bm{\mathcal{E}} \cdot \bm{\mathcal{S}} &= 1 \cdot 0.3 + 0 \cdot 0.1 + 0 \cdot 0.5 + 0 \cdot 0.1 \\
&= 0.3
\end{align*}
This score reflects the total attribution the method assigns to the known evidence (``Karla''). A higher dot product indicates better alignment with known evidence.

\paragraph{Average Precision (AP):}
This metric evaluates how well the attribution method ranks the evidence token(s). We start by sorting the attribution scores in descending order and recording their indices:

\[
\text{Rank}(\bm{\mathcal{S}}) = [2, 0, 1, 3]
\]
This means the token at index 2 (``about'') has the highest score, followed by index 0 (``Karla''), and so on. The known evidence index (based on $\bm{\mathcal{E}}$) is:
\[
\text{Evidence Index} = \{0\}
\]

To compute AP, we iterate through the ranked list and monitor when the evidence index is included. Precision is calculated at each step where recall increases (i.e., when a new evidence token is found):

\begin{itemize}
  \item [1.] \textbf{[2]} — does not include index 0 → no change in recall
  \item [2.] \textbf{[2, 0]} — includes index 0 → recall changes → precision = 1/2 = 0.5
  \item [3.] \textbf{[2, 0, 1]} — no change in recall
  \item [4.] \textbf{[2, 0, 1, 3]} — no change in recall
\end{itemize}
So, the final AP is:
\[
\text{AP} = 0.5
\]

Note that if the evidence index were 2 (evidence ranked first), the AP would be $1.0$. If it were 1 (ranked third), the AP would be approximately $0.33$. A higher AP indicates better alignment between the attribution method’s ranking and the true evidence.

\end{document}